\documentclass[Afour,sageh,times]{sagej}

\usepackage{moreverb,url}

\usepackage[colorlinks,bookmarksopen,bookmarksnumbered,citecolor=red,urlcolor=red]{hyperref}
\usepackage[utf8]{inputenc}
\usepackage{multicol}
\usepackage{textcomp}
\usepackage{subcaption}
\usepackage{amsmath}
\usepackage{booktabs} % For better table lines
\usepackage[title]{appendix}%
\usepackage{adjustbox} % For adjusting table size
\usepackage{pgfplots}
\pgfplotsset{compat=1.17}
\usepackage{pgfplotstable}
\usepackage{geometry}
\usepackage{float}
\usepackage[absolute,overlay]{textpos} 
\usepackage{ragged2e}
\usepackage{graphicx}      % For including graphics  % For subfigures
\usepackage{siunitx}
\usepackage{picture}       % For annotations
\usepackage{placeins} 
\usepackage{lastpage}
\usepackage{etoolbox}
\newcommand\BibTeX{{\rmfamily B\kern-.05em \textsc{i\kern-.025em b}\kern-.08em
T\kern-.1667em\lower.7ex\hbox{E}\kern-.125emX}}

\makeatletter
\patchcmd{\@maketitle}
  {\raggedright\titlesize\textbf{\@title}}
  {\centering\titlesize\textbf{\@title}}
  {}{}
\patchcmd{\@maketitle}
  {\raggedright\textbf{\@author}}
  {\centering\textbf{\@author}}
  {}{}
\makeatother

\begin{document}

% \runninghead{Ansari et al.}

\title{Hybrid Robotic Meta-gripper for Tomato Harvesting: Analysis of Auxetic Structures with Lattice Orientation Variations}

\author{Shahid Ansari\affilnum{1}, Vivek Gupta\affilnum{2} and Bishakh Bhattacharya\affilnum{3}}

\affiliation{\affilnum{1}Tohoku University, Japan\\
\affilnum{2}Delft University of Technology (TU Delft), the Netherlands\\
\affilnum{3}Indian Institute of Technology Kanpur (IIT Kanpur), India}

\corrauth{Shahid Ansari,
Space Robotics Laboratory,
Tohoku University, Japan}

\email{ansari.shahid.b2@tohoku.ac.jp}

\begin{abstract}
The agricultural sector is rapidly evolving to meet growing global food demands, yet tasks like fruit and vegetable handling remain labor-intensive, causing inefficiencies and post-harvest losses. Automation, particularly selective harvesting, offers a viable solution, with soft robotics emerging as a key enabler.
This study introduces a novel hybrid gripper for tomato harvesting, incorporating a rigid outer frame with a soft auxetic internal lattice. The six-finger, 3D caging-effect design enables gentle yet secure grasping in unstructured environments. Uniquely, the work investigates the effect of auxetic lattice orientation on grasping conformability, combining experimental validation with 2D Digital Image Correlation (DIC) and nonlinear finite element analysis (FEA). Auxetic configurations with unit cell inclinations of 0°, 30°, 45°, and 60° are evaluated, and their grasping forces, deformation responses, and motor torque requirements are systematically compared. Results demonstrate that lattice orientation strongly influences compliance, contact forces, and energy efficiency, with distinct advantages across configurations. This comparative framework highlights the novelty of tailoring auxetic geometries to optimize robotic gripper performance. The findings provide new insights into soft–rigid hybrid gripper design, advancing automation strategies for precision agriculture while minimizing crop damage.

\end{abstract}

\keywords{Precision Agriculture, Robotic Gripper, Metamaterial, Digital Image Correlation(DIC)}

\maketitle

\section{Introduction}
The agriculture sector plays a pivotal role in addressing global challenges such as labor shortages and food security. Automation has emerged as a crucial tool to enhance efficiency and sustainability in agriculture (\cite{sistler1987robotics, horn1999agriculture}). Among agricultural processes, harvesting is particularly labor-intensive and critical, especially for delicate crops like tomatoes (\cite{onishi2019automated, bac2014harvesting}). Tomatoes are prone to damage during the picking process, making it challenging for conventional harvesting methods to balance operational efficiency with the preservation of product quality (\cite{li2011review}). These challenges have spurred interest in compliant, shape-adaptive robotic systems capable of performing gentle and precise harvesting operations *(\cite{elfferich2022soft}).
Existing automated mechanical graspers for tomato harvesting often struggle with inconsistent grasping forces and limited adaptability to variations in fruit size and ripeness levels (\cite{zhang2018comparative, kaur2019toward}). Recent advancements in robotic gripper designs, particularly those leveraging metamaterials, have demonstrated remarkable potential in improving structural adaptability and durability. Lattice metastructures, such as hourglass-shaped designs (\cite{gupta2020exploring, gupta2022energy, gupta2024evidence}) which are known for customizable stiffness and negative stiffness metamaterials (\cite{dwivedi2020simultaneous}), are being integrated into soft robotic systems to enhance resilience and grasping efficiency. Furthermore, novel approaches like voxel-scale metamaterial topology optimization have been applied to customize soft grippers for specific agricultural needs (\cite{pinskier2024towards}).

Metamaterial-based grippers have proven particularly effective in handling fragile crops, leveraging their unique structural properties to achieve conformal and damage-free grasping (\cite{guzman2024metamaterial, zhu2019fully}). The integration of 3D-printed metamaterials with mechano-optic force sensors has enabled precise force control in agricultural applications, highlighting the role of programmable structures in advanced robotic systems (\cite{hegde20243d, berwind2018hierarchical}). Additionally, hybrid designs that combine soft actuators with rigid, metamaterial-inspired skeletons have shown promising results in adjusting mechanical properties for varying harvesting conditions (\cite{zhang2024hybrid}).

Machine learning and computational optimization techniques are further augmenting the design process of robotic grippers. Deep reinforcement learning, for instance, has been employed to optimize metamaterial mechanisms for compliance control (\cite{choi2024deep}), while computational multi-sensor systems are being used to enhance grip-strength adaptability (\cite{chen2023metamaterial}). Continuum robot arms equipped with specialized grippers have demonstrated effectiveness in navigating complex environments to harvest tomatoes and similar crops (\cite{kultongkham2021design, yeshmukhametov2022development}). The role of sensory feedback in robotic grippers, as explored in designs with force sensors and adaptive feedback loops, has also been critical in minimizing fruit damage during harvesting (\cite{ansari2025design, cacucciolo2019delicate}).

This paper introduces a novel hybrid robotic gripper specifically designed for tomato harvesting. The gripper employs a re-entrant honeycomb auxetic structure—a type of metamaterial—integrated with a rigid link mechanism to enable gentle fruit handling while maintaining a firm grasp. The design also incorporates force sensors and an advanced control system to enhance adaptability and precision, minimizing fruit damage and improving operational performance.
The proposed gripper system features a hybrid design combining soft, compliant materials with rigid structures to achieve the flexibility required for delicate handling and the strength needed for robust grasping. Each of the six rigid-link fingers includes an internal re-entrant honeycomb auxetic structure, connected through wedge-shaped ports. The gripper’s performance was evaluated through extensive experimental trials, including 2D Digital Image Correlation (DIC) analysis of auxetic structure deformation, finite element analysis, and motor torque estimation for various auxetic configurations. Metrics such as grasping force, fruit damage rates, and harvesting efficiency were analyzed, demonstrating a significant reduction in fruit damage compared to conventional methods.

This research represents a significant advancement in the development of automated, high-precision harvesting systems. The remainder of this paper is structured as follows: Section 2 describes the design methodology of the proposed gripper. Section 3 discusses the experimental setup and results, and Section 4 concludes with key findings and future directions.
\begin{table*}[t]
\small\sf\centering
\caption{Overview of Grippers Utilizing Metamaterials and Auxetic Structures for Enhanced Grasping Applications.\label{tab:Overview_Metagrippers}}
\setlength{\tabcolsep}{4pt}
\begin{tabular}{@{}p{0.46\textwidth} p{0.44\textwidth} p{0.10\textwidth}@{}}
\toprule
\textbf{Gripper Mechanism} & \textbf{Application} & \textbf{Reference} \\
\midrule
3D-printed modular soft gripper with integrated auxetic metamaterial & Conformal grasping of diverse objects & \cite{tawk2021} \\
Soft gripper with auxetic metamaterial mesh integrated on pneumatic fingers & Gripping delicate or multiply curved surfaces & \cite{US20210016452A1} \\
Shape-memory alloy-based auxetic smart metamaterial applied to a four-jawed gripper & Adaptive gripping with enhanced functionality & \cite{wang2024} \\
Electrically driven soft robotic gripper using handed shearing auxetics & Energy-efficient gripping with high deformation & \cite{hawkes2019} \\
Hybrid soft electrostatic metamaterial gripper & Multi-surface, multi-object adaptation & \cite{kanno2024} \\
Programmable soft bending actuators with auxetic metamaterials & Soft robotic applications requiring programmable bending & \cite{pan2020} \\
Vacuum-driven auxetic switching structure applied to a gripper & Adaptive gripping with vacuum actuation & \cite{liu2020} \\
Smart compliant robotic gripper with 3D-designed cellular fingers & Enhanced adaptability and compliance in grasping & \cite{kaur2019} \\
Mechanical metamaterials for sensor and actuator applications & Enhancing sensor and actuator performance & \cite{wang2023} \\
Development, fabrication, and mechanical characterization of auxetic structures & Designing auxetic structures for various applications & \cite{smith2023} \\
\bottomrule
\end{tabular}
\end{table*}

\section{Methodology}
\subsection{Experimental section}
\subsubsection{Design of 3D printed Hybrid Gripper}
\begin{figure*}[!htbp]
\centerline{\includegraphics[width=1.5\columnwidth]{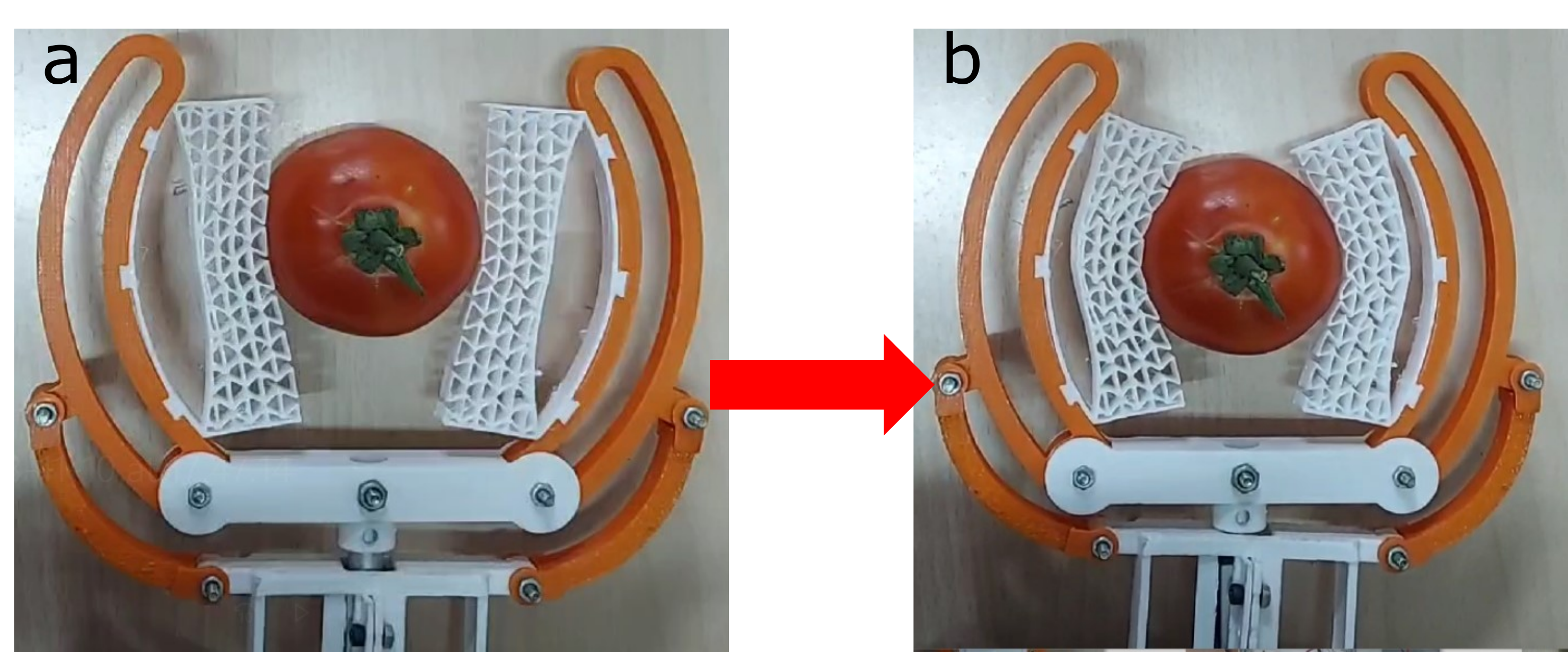}}
\caption{Two-dimensional view of auxetic structure interaction with the tomato to have a high shape conformability when the rigid link mechanism of the gripper actuates (Fig. 2(a): open position and Fig. 2(b): closed position) }
\label{fig:auxetic_deformation}
\end{figure*}

In this paper, a hybrid gripper design is introduced, which combines rigid and compliant elements. The gripper features curved rigid linkages and soft re-entrant passive auxetic structures, providing excellent shape adaptability for grasping. With the use of six fingers, the gripper offers a strong caging effect, securely holding the tomato in place without causing any damage. Each passive, soft structure comprises a 2D re-entrant auxetic pattern supported by an arch-shaped leaf spring. This configuration allows for shape conformance to the surface of the tomato, as illustrated in Fig.\ref{fig:auxetic_deformation}\par
  
The figure shows a rigid link mechanism that has been 3D printed using PLA (polylactic acid), known for its strength. In contrast, the soft auxetic structures are 3D printed using commercially available soft thermoplastic polyurethane (TPU) material. The geometric properties are provided in Table \ref{tab:geometric_properties} while the material properties can be seen in Table \ref{tab:material_properties} present in the appendix, respectively.\par
  
\begin{figure*}[!htbp]
\centerline{\includegraphics[width=1.5\columnwidth]{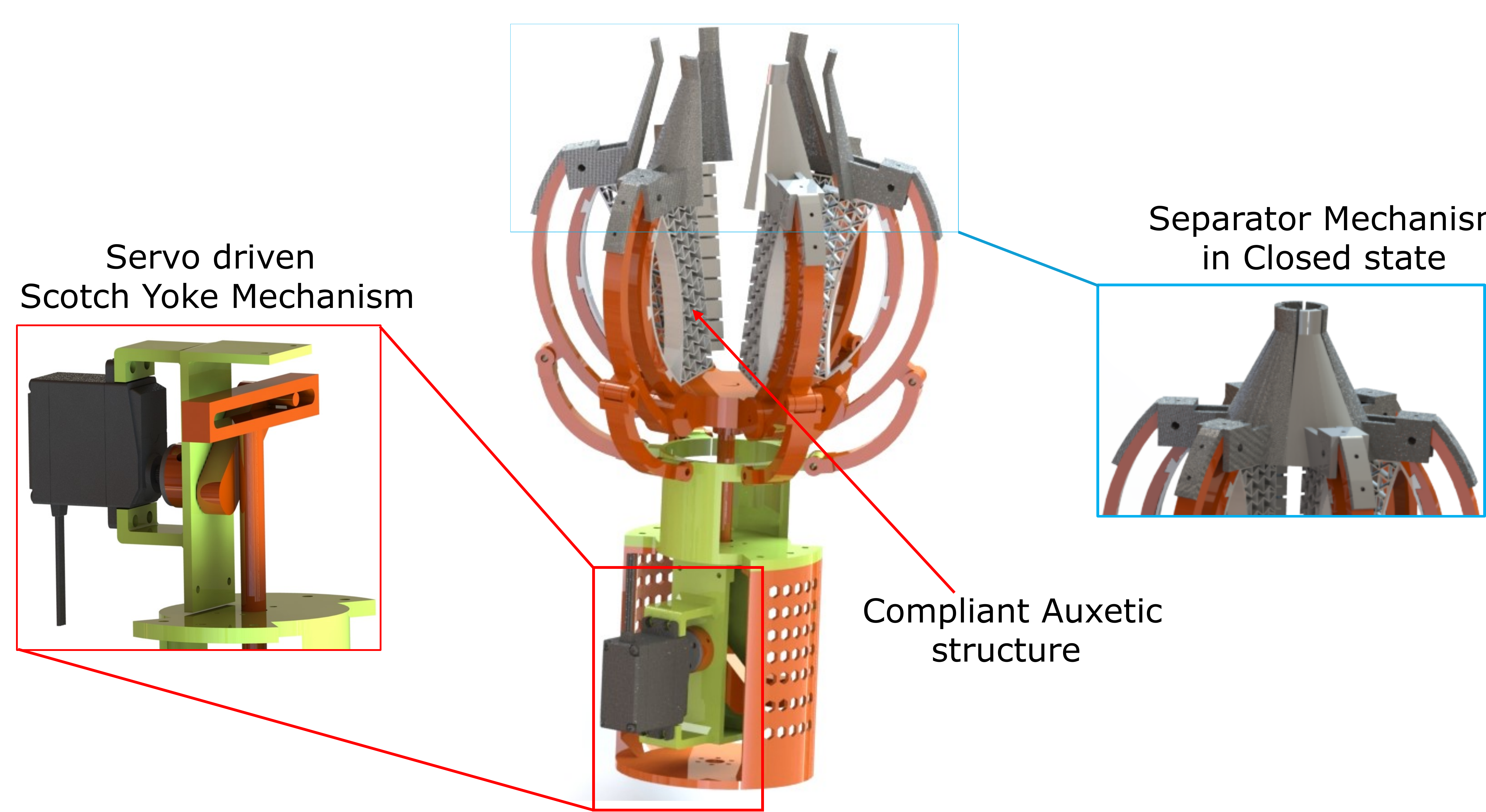}}
\caption{CAD model of the proposed gripper with various parts: the outer rigid link mechanism attached with soft auxetic structures. The linkage drive is shown on the left of the figure, while the tomato separator mechanism is shown on the right side of the figure. }
\label{fig: Gripper_CAD_Assembly}
\end{figure*}

 The gripper's mechanism operates by moving its individual digits through the fourth inversion of the slider-crank mechanism. Six of these units are arranged in six different planes, each 60$^{\circ}$ apart, and are connected to create a cage-like structure. This structure is controlled by a single Scotch yoke mechanism powered by a servo motor. The servo motor is linked to the Scotch yoke mechanism through a crank, as shown in the inset of Fig. \ref{fig: Gripper_CAD_Assembly}. The plunger is connected to the sliding shaft of the Scotch yoke mechanism to operate the gripper's opening and closing. Some parts of the Scotch yoke mechanism are fixed inside a rectangular slot on the support structure. To prevent the tomato from falling out, a flexible latex basket that adheres to the auxetic fingers is used to constrain it.
% \subsection{Design of 3D printed Auxetic structures}
\subsubsection{Mechanical characterization of auxetic structure designs using Digital Image Correlation(DIC)}
\begin{figure*}[!htbp]
      \centering
      \includegraphics[width=0.8\linewidth]{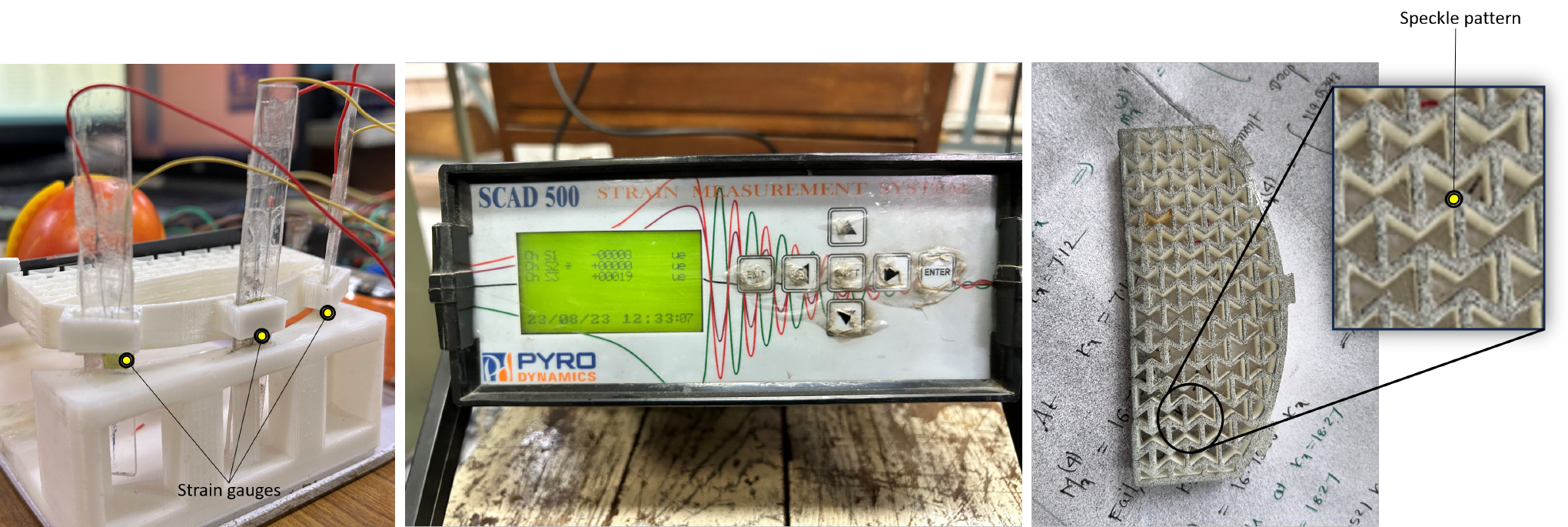}
       \caption{The figure shows the various parts of the experimental setup, which are as follows: The left-hand side of the figure shows the strain gauges mounted on the 3D printed structure, the middle of the figure shows the strain measurement system, while the right-hand side of the figure shows the speckle pattern created through paint spray on the surface of the auxetic structure to improve the accuracy of the image correlation}
      \label{fig:strain_gauge_system}
\end{figure*}

To experimentally determine the level of force while an auxetic structure-based soft finger interacts with the tomato and the reaction forces that it encounters at the connecting ports on the rigid links of the gripper mechanism, an experimental setup was fabricated by fused deposition modeling (FDM) using white PLA (polylactic acid) filament. The body of this setup is mounted with three acrylic cantilever beams and a strain gauge mounted on each of them. The specifications of these strain gauges can be depicted from the table \ref{tab:strain_gauge_specification}. Each of the three connecting ports of the auxetic structure can be attached to the beam through a 3D-printed connector fabricated using PLA filament, which can be depicted in the \ref{fig:strain_gauge_system}. With the application of force on the auxetic structure contact surface, we can directly measure the reaction forces on each of these ports through strain gauges. The complete procedure of the reaction force measurement is explained in detail in appendix \ref{appendix:strain_measurement_process} where the forces of reaction generated during the tomato interaction with the auxetic structure can be visualized in \ref{fig:Tomato_interaction_reaction forces} and the theory of cantilever beam used for determining the value of reaction loads can be depicted in \ref{fig:Cantilver_beam_strainggauge}. These strain gauges are connected to the three-quarter bridge channels of SCAD 500 Strain Measurement System as can be depicted in the middle of the figure \ref{fig:strain_gauge_system}, which is an advanced device by the Pyrodynamics company and used in experimental stress analysis and mechanics. The SCAD 500 is designed with two amplified transducer inputs and incorporates the latest digital circuitry, making it fully programmable. It offers an online display of data in ASCII format and has selectable gain options. The system can accept any combination of quarter, half, or full bridge strain gauge circuits: 120 Ohms, 350 Ohms, and 1000 Ohms. The device boasts excellent overall accuracy and stability. The digital display on the device provides real-time data, aiding in precise and accurate measurements (\cite{pyrodynamics_products_services}).
The real-time data from these strain gauges can be captured through data acquisition by NI DAQ, which is a National Instruments BNC-2110 connector block data acquisition system (\cite{ni_bnc_2110}) that will display the real-time data from the strain gauges on the computer through a Labview interface, which is a Laboratory Virtual Instrument Engineering Workbench is a type of system-design platform and development environment for a visual programming language developed by National Instruments. 
The contact-making face of the auxetic structure with the tomato is installed with a long FSR sensor strip, an RP-L-170 thin film flexible Pressure sensor with a length of 170 mm, designed for susceptible pressure detection. It is made of ultra-thin film of excellent mechanical properties, conductive materials, and nanometer pressure-sensitive layers. The sensor can sense static and dynamic pressure with a high response speed. The detailed specifications of this sensor can be found in \cite{dfrobot_rp_l_170}
To test the capabilities of the auxetic structure, a tomato is placed inside a 3D printed supporting holder, which is driven by a linear actuator connected to it; the actuator we have used is L16-P Miniature Linear Actuator with Feedback 140 mm 63:1 12 volts by Actuonix company (\cite{actuonix_l16_140_63_12_p}). The actuation is performed with the help of the Atmega328P microcontroller board. When the linear actuator pushes the tomato towards the surface of the auxetic structure, it undergoes deformation, which is captured by a DIC camera, which captures the fine details of the deformation that occurs in the complete speckle pattern over the auxetic front face, which can be depicted from the left-hand side of the figure \ref{fig:strain_gauge_system}.
Simultaneously, the real-time data of the contact force and the reaction forces at various instants of time through the FSR sensor and three strain gauges is captured. The data from the FSR sensor is linked to a data acquisition system based on the Atmega328P microcontroller. This system collects contact force data, which can be stored directly in PLX-DAQ, which is an add-on tool by Parallax for Microsoft Excel (\cite{parallax_plx_daq}). It allows any microcontroller connected to a sensor and a PC's serial port to transmit data directly to Excel and aids in further post-processing.

\begin{figure*}[!htbp]
    \centering
    \includegraphics[width=0.8\linewidth]{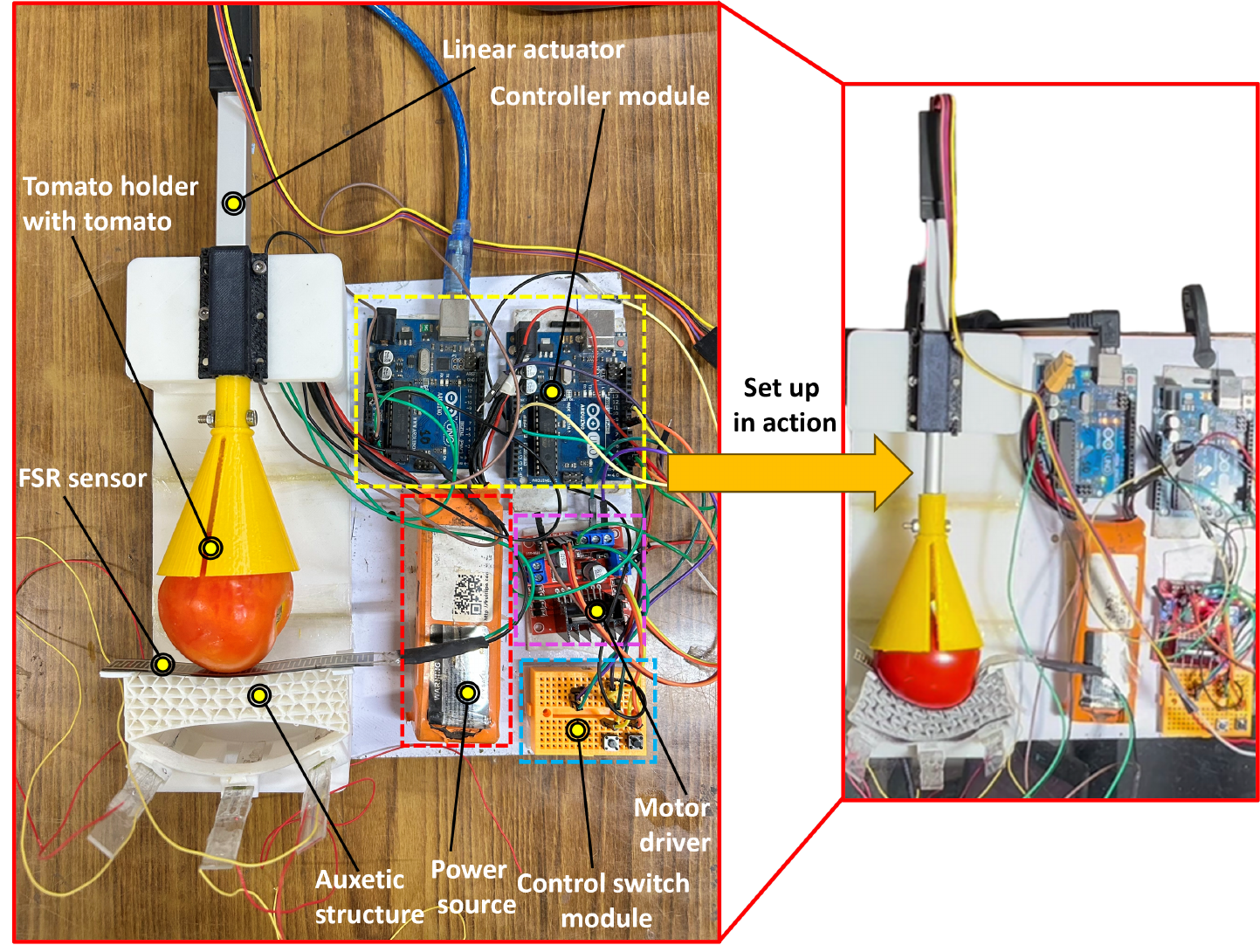}
    \includegraphics[width=0.8\linewidth]{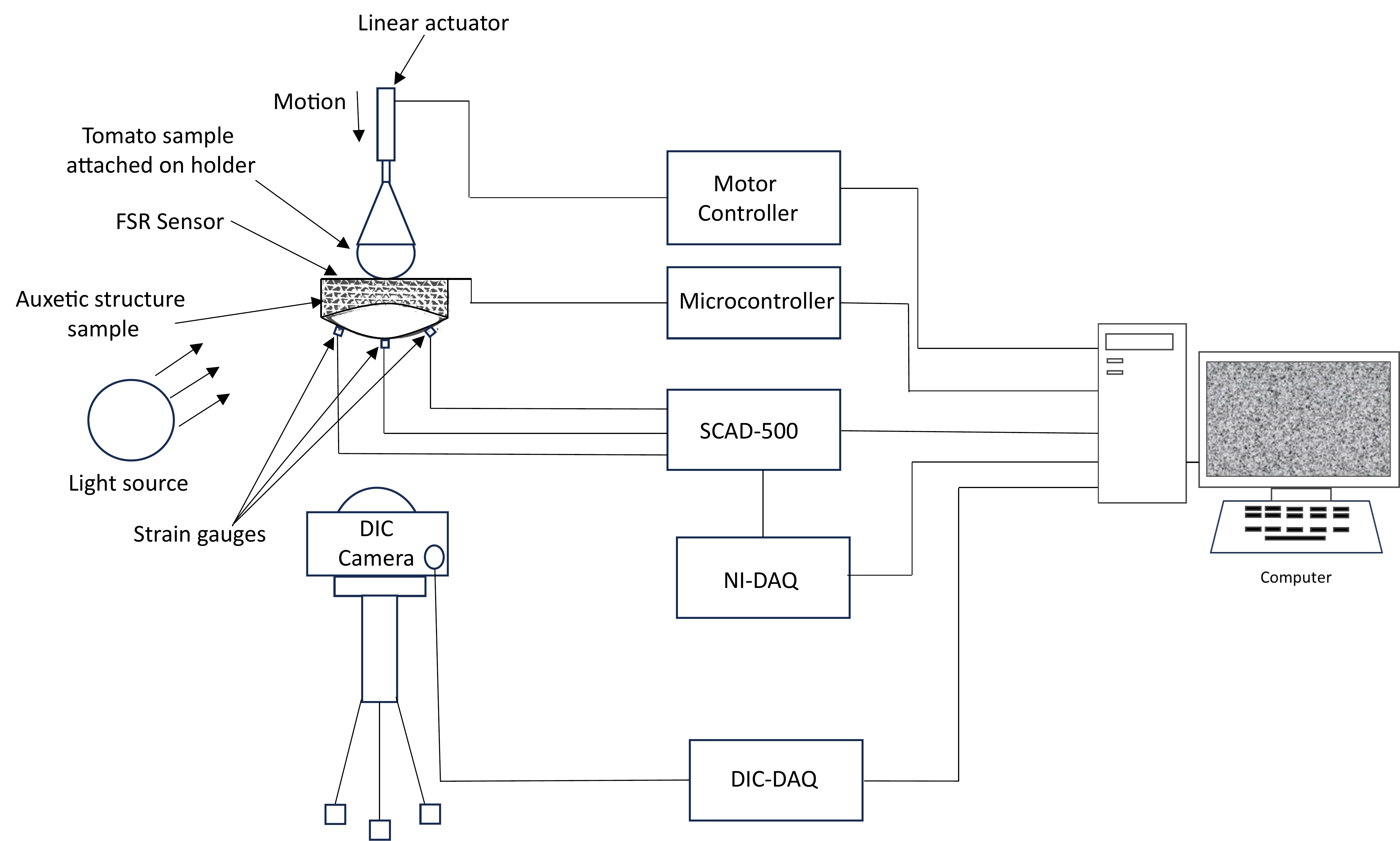}
    \caption{Experimental setup inclusive all mechatronics components used for analyzing auxetic structure samples. The left-hand side shows the various components of the setup designed to measure interaction forces, while the right-hand side depicts the setup in action. This setup facilitates the collection of multimodal information for a comprehensive analysis.}

    \label{fig:Experimental_setup and schematic}
\end{figure*}

\begin{figure*}[!htbp]
    \centering
    \includegraphics[width=0.8\linewidth]{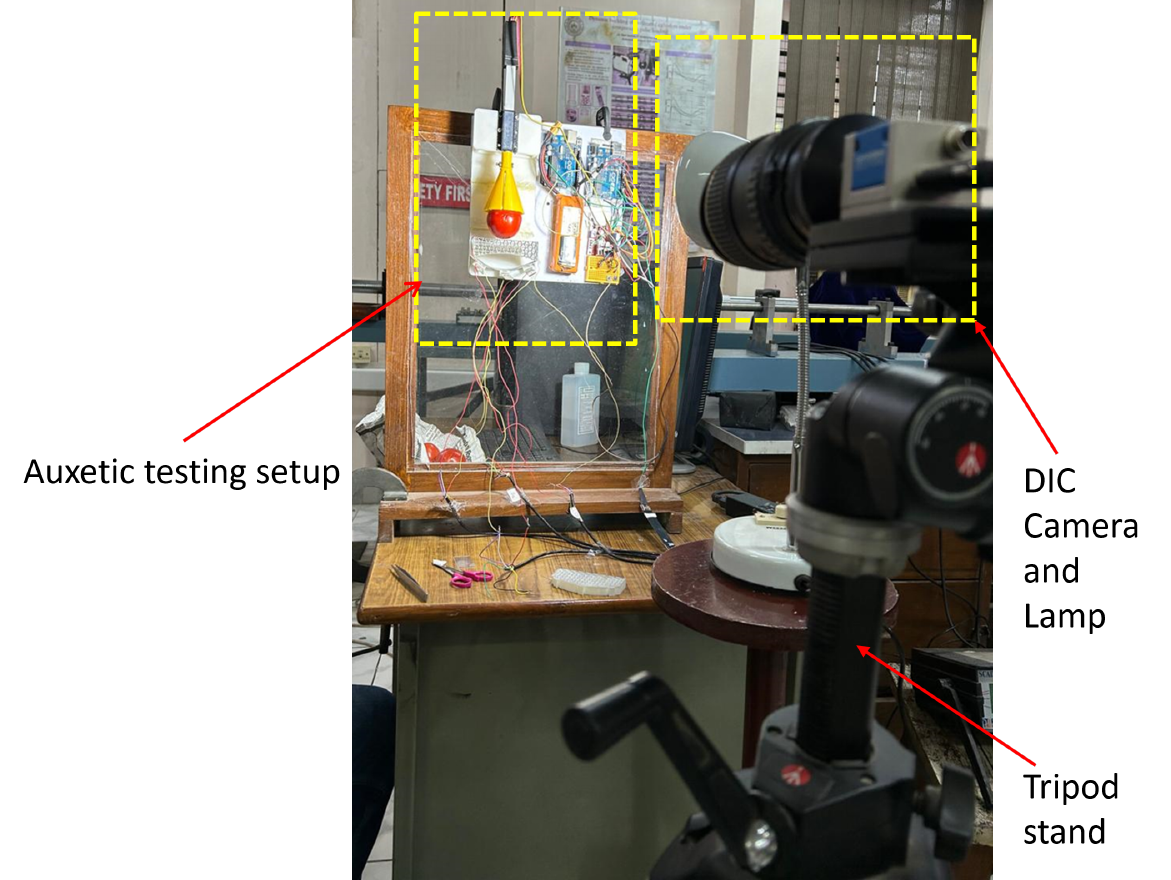}
    \includegraphics[width=0.7\linewidth]{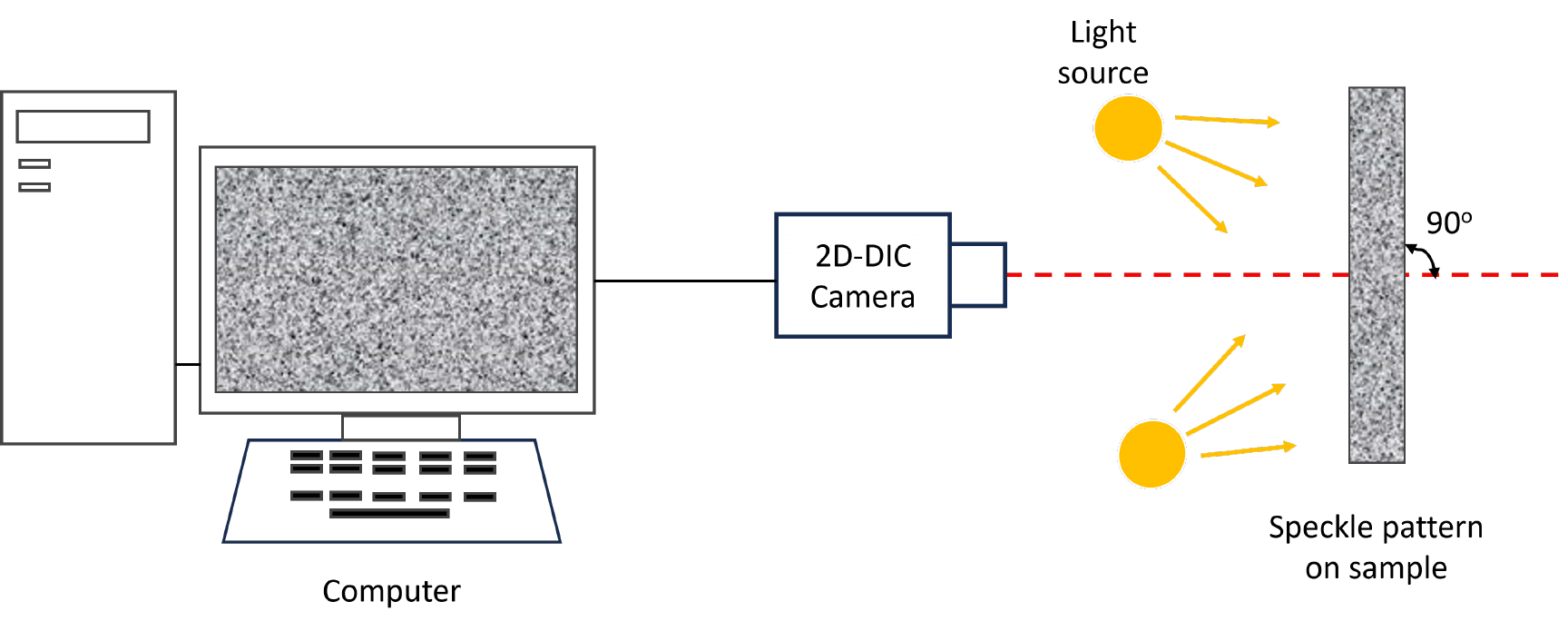}
    \caption{The figure illustrates the complete setup and schematic diagram used for the digital image correlation process in the experimental deformation analysis of auxetic structure samples.}

    \label{DIC_camera_process_schematic}
\end{figure*}

To measure deformation and strain distribution inside the complete auxetic structure as can be depicted from the figure \ref{fig:Experimental_setup and schematic}, we have performed 2D DIC (Digital Image Correlation) using a camera by Correlated Solutions, a leading provider of Digital Image Correlation (DIC) systems, which can be depicted from the figure \ref{DIC_camera_process_schematic}. This camera is likely a component of a turnkey solution designed to deliver full-field, 2D displacement and strain data for mechanical testing on planar specimens. Correlated Solutions DIC systems typically incorporate high-resolution monochrome cameras. These cameras offer resolutions ranging from 2.3 to 31 Megapixels for quasi-static testing systems, and for high-speed testing, they include high-speed monochrome digital cameras with frame rates up to 40,000 fps at full resolution. The camera's compatibility with a range of hardware and its adaptability to various testing requirements make it a versatile tool in the field of mechanical testing.

A dedicated software interface, VIC-2D, is used in this case by Correlated Solutions (\cite{correlatedsolutions_vic2d}). The Video Image Correlation in 2D (VIC-2D) system is a non-contact, full-field, two-dimensional displacement and strain data provider for mechanical testing on planar specimens. It employs optimized correlation algorithms to measure in-plane displacements at every point within the area of interest. The full-field strain is computed and displayed to identify strain concentrations easily. The schematic diagram of the DIC process performed for analyzing the deformation of the proposed auxetic structure samples having speckle pattern on its surface can be depicted from the figure \ref{DIC_camera_process_schematic}. With extremely low noise levels and high accuracy, sub-micron displacements can be measured, and local strains from 50 microstrains to over 2500\% are achievable. The system can measure specimen sizes ranging from microscopic to very large scale by simply adjusting optics.
\section{Results and Discussion}
\subsection{Structural Analysis of Auxetic Structure Samples through 2D DIC}

Since we have proposed the idea of using the re-entrant honeycomb auxetic structures with negative Poisson's ratio for grasping the tomatoes delicately, it is quite essential that the system should have high shape conformability. Based on the experimental analysis using Digital Image Correlation (DIC), we have analyzed the auxetic structures through two parameters, i.e., strain in the vertical direction ($e_{yy}$), and the vertical displacement $v$ in the direction of loading when the tomato interacts with it during testing.

\footnote{Note: To simplify the discussion, we have denoted the auxetic structure samples as Aux (unit cell inclination with the vertical axis). Hence, for four different auxetic structure samples, the representation will be Aux($0^{\circ}$), Aux($30^{\circ}$), Aux($45^{\circ}$), Aux($60^{\circ}$) respectively.}

\begin{figure*}[!htbp]
\centerline{\includegraphics[width=1.5\columnwidth]{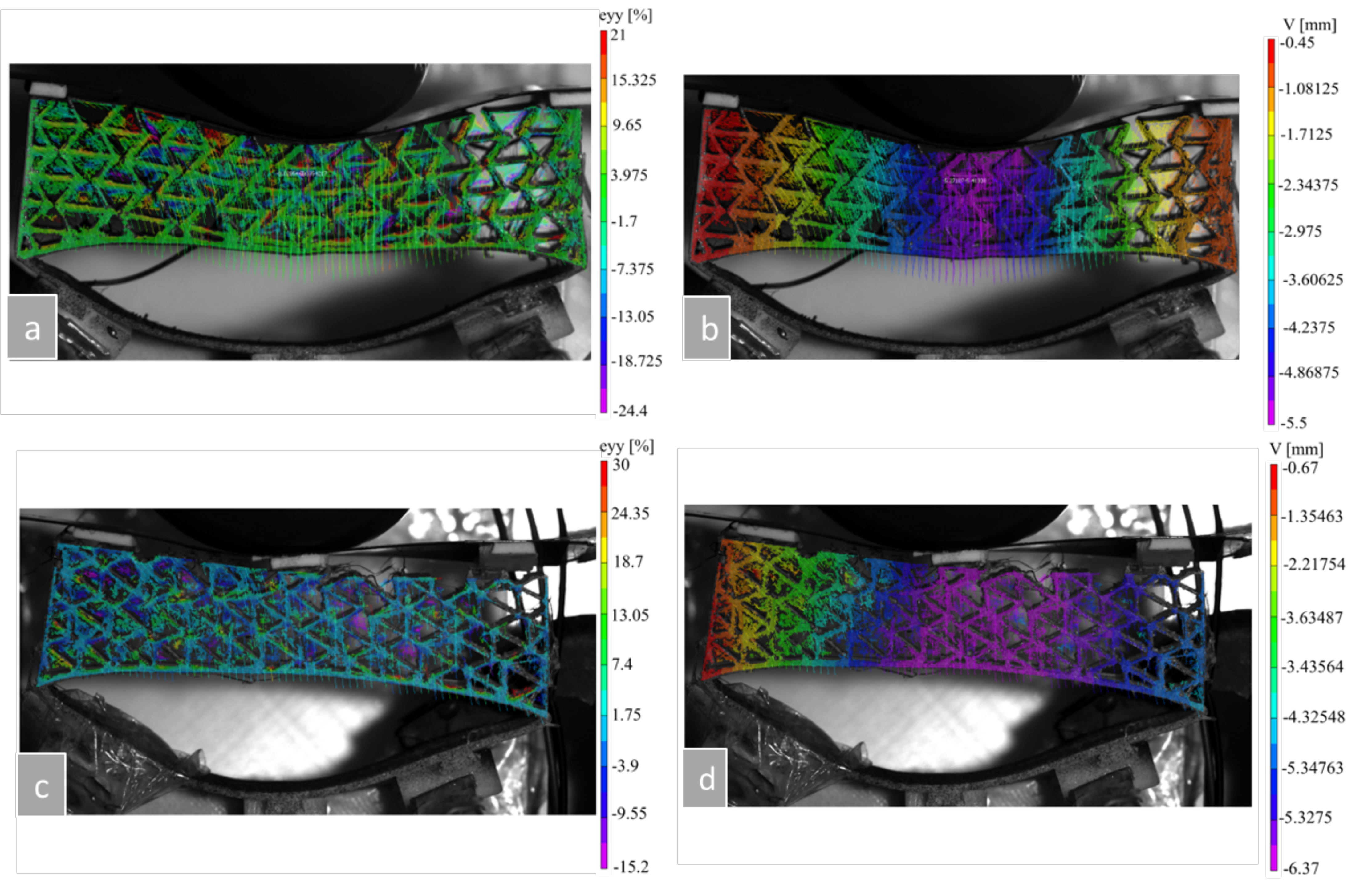}}
\centerline{\includegraphics[width=1.5\columnwidth]{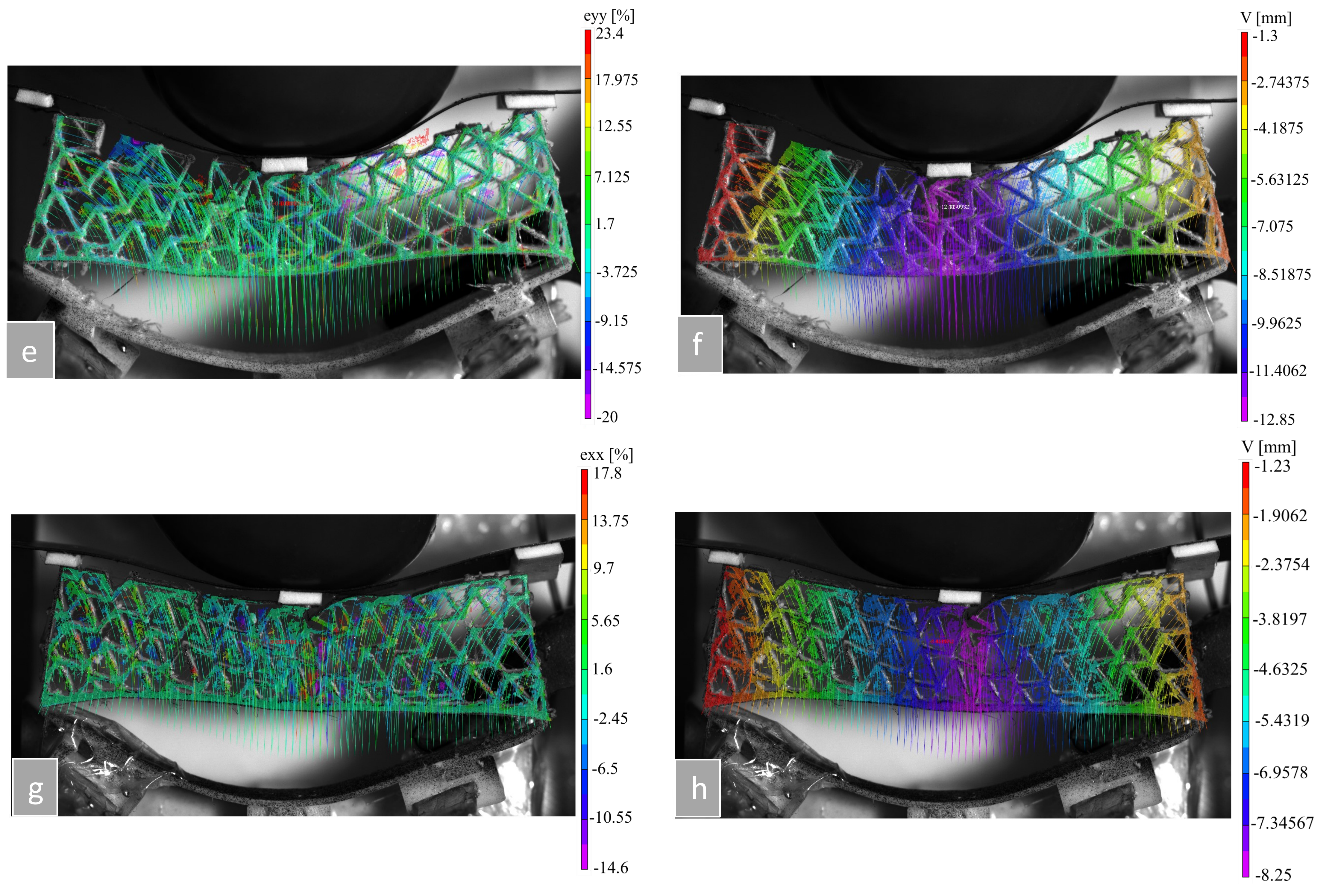}}
\caption{The figure illustrates the variation of strain and deflection in the y-direction along a horizontal central line of the auxetic structure for different unit cell inclinations with respect to the x-direction. Subfigures (a) \& (b) represent these variations for the $0^{\circ}$ unit cell inclination, (c) \& (d) for the $30^{\circ}$ unit cell inclination, (e) \& (f) for the $45^{\circ}$ unit cell inclination, and (g) \& (h) for the $60^{\circ}$ unit cell inclination. These plots highlight the deformation behavior of the auxetic structures under different inclinations.}
\label{fig: DIC_Colormaps}
\end{figure*}

\begin{figure*}[!htbp]
    \centering
    \includegraphics[width=\textwidth]{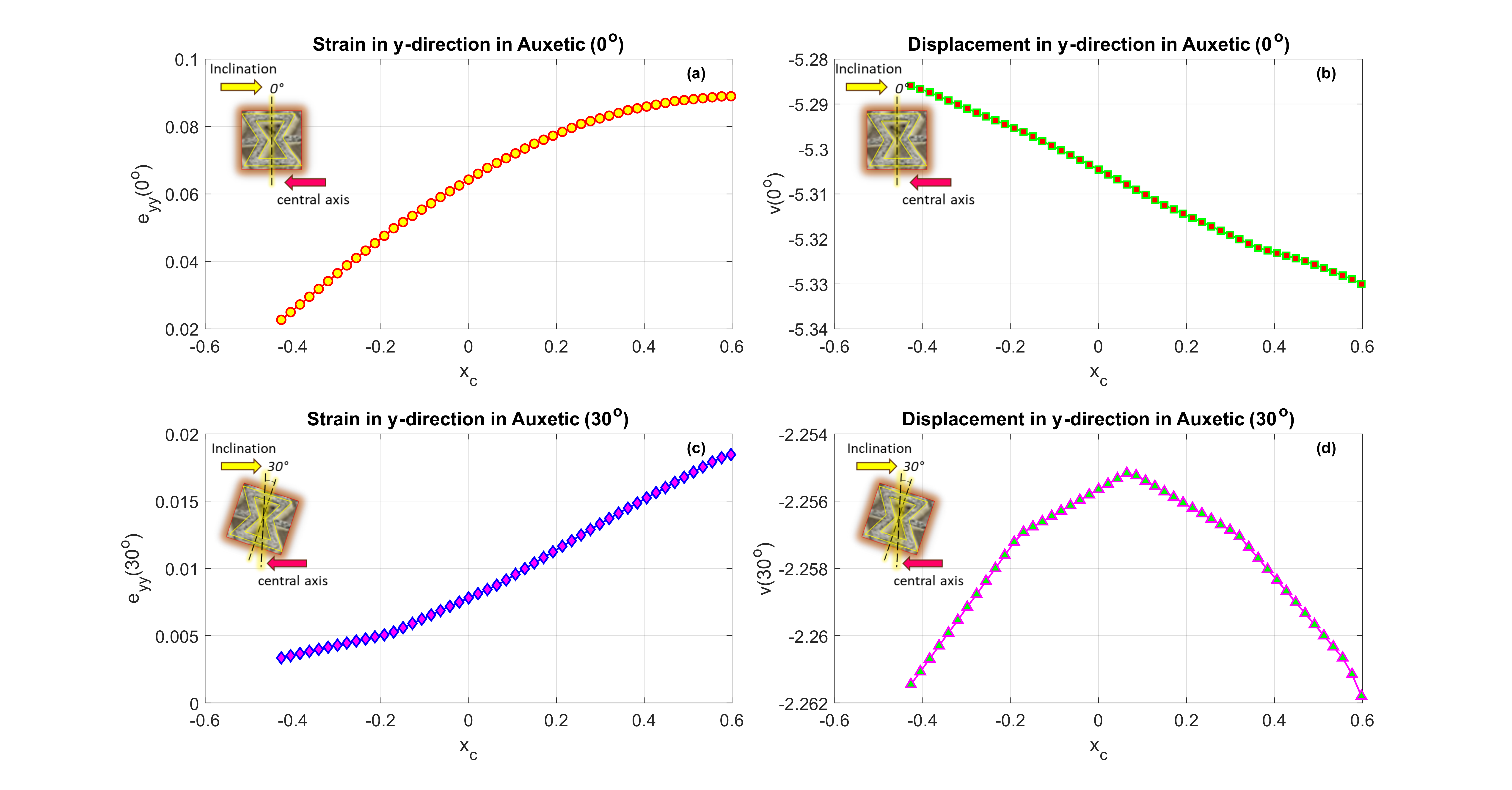}  \includegraphics[width=\textwidth]{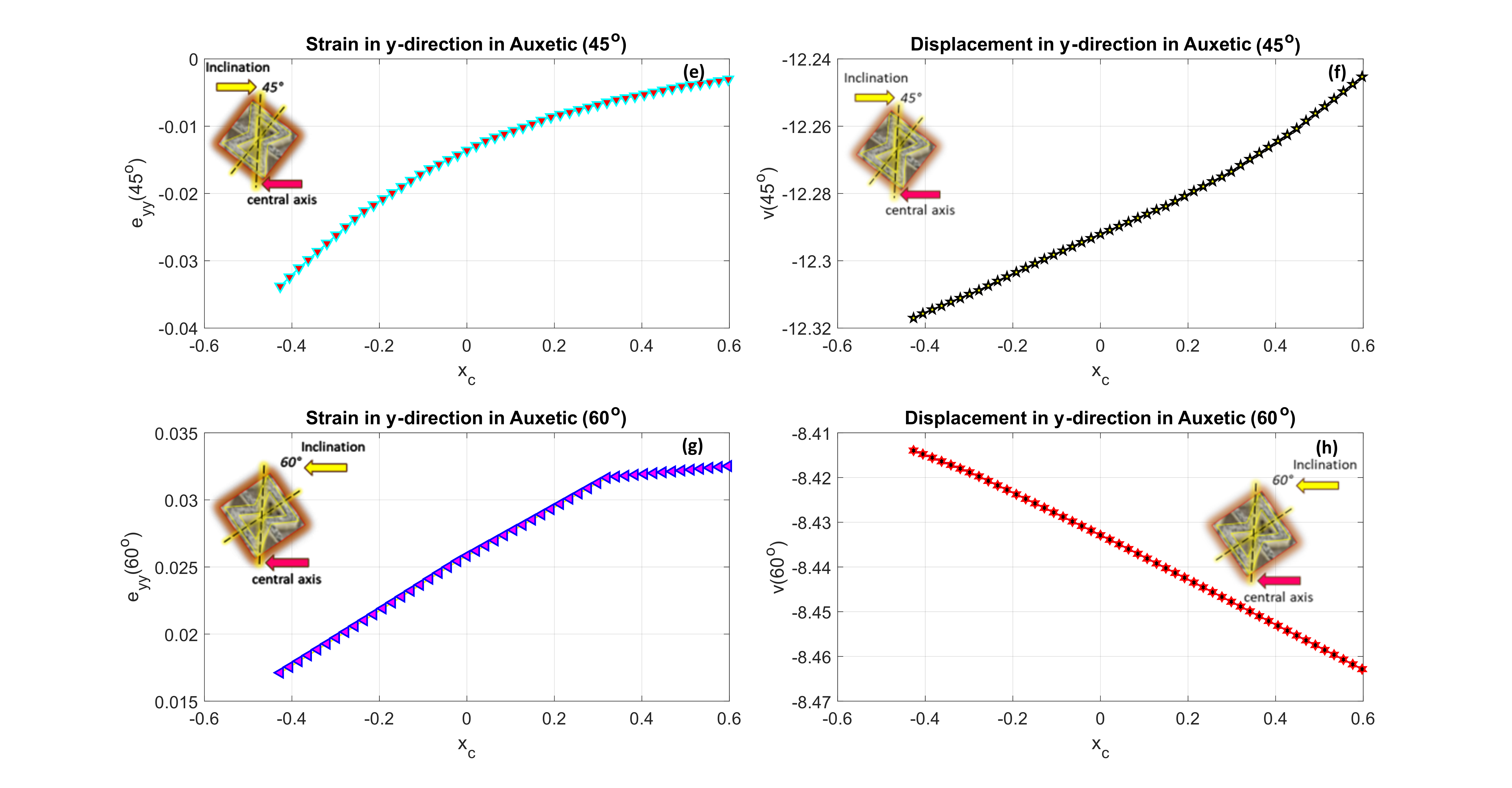}
    \caption{Plots illustrating the mechanical response of auxetic structure samples under applied load. Subfigures (a), (c), (e), and (g) present the strain-displacement plots for Aux($0^{\circ}$), Aux($30^{\circ}$), Aux($45^{\circ}$), and Aux($60^{\circ}$), showing how strain varies with respect to the x-direction and highlighting the deformation phase. Subfigures (b), (d), (f), and (h) depict the variation of displacement with respect to the x-direction for the same auxetic structure samples, providing further insights into their deformation behavior. Together, these plots capture the mechanical response and deformation phases of the auxetic structure samples.}
    \label{fig:strain_displacment_plots_matlab}
\end{figure*}

From our analysis based on the data collected, we have found that the auxetic structure Aux($45^{\circ}$) exhibited superior performance compared to other samples of auxetic structures. Its strain of 0.11 and displacement of 12.50 mm (variation can be seen in fig.\ref{fig:strain_displacment_plots_matlab}(a) and (b) respectively), significantly outperformed the values for auxetic structure Aux($0^{\circ}$) (with strain: 0.09, displacement: 5.29 mm) (Variation can be seen in fig.\ref{fig:strain_displacment_plots_matlab} (c) and (d) respectively), for auxetic structure Aux($30^{\circ}$) (with strain: 0.02, displacement: 2.28 mm) (Variation can be seen in fig.\ref{fig:strain_displacment_plots_matlab} (e) and (f) respectively), while for auxetic structure Aux($60^{\circ}$) (with strain: 0.03, displacement: 8.49 mm) (Variation can be seen in fig.\ref{fig:strain_displacment_plots_matlab} (g) and (h)). The strain and displacement distribution of auxetic structure samples Aux($0^{\circ}$), Aux($30^{\circ}$), Aux($45^{\circ}$) and Aux($60^{\circ}$) can be depicted from the figure \ref{fig: DIC_Colormaps} (a, b), figure \ref{fig: DIC_Colormaps} (c, d), figure \ref{fig: DIC_Colormaps} (e, f) and figure \ref{fig: DIC_Colormaps} (g, h) respectively.
%%%%%Colormaps
%%%Strain displacement matlab plots
\subsection{Analysis of curvature of auxetic structures}

To examine how curvature varies along the contact surface in the x-direction, we determine the local curvature \( k \) at each point on the surface profile, given by \( y = f(x) \). The curvature \( k \) reflects the rate at which the tangent angle changes, providing information on the local geometric characteristics of the contact interface:

\[
k = \frac{|y''|}{(1 + (y')^2)^{3/2}}
\]

where $y' = \frac{dy}{dx}$ is the slope and $y'' = \frac{d^2y}{dx^2}$ represents the rate of change of the slope. This formula considers both the slope and its variation at each point, yielding an accurate measure of the curvature along $x$.

By plotting \( k \) as a function of \( x \), we generate a curvature profile that illustrates how the contact surface geometry adjusts during interaction. This curvature analysis is essential for comprehending the deformation behavior of auxetic structures, especially in applications that demand adaptable contact interfaces.

To quantitatively perform the comparative study of the curvature of the four samples of the auxetic structures to determine the best candidate, the important point is to consider the influence of curvature on the gripper's ability to conform to the shape of the soft fruit. In general, the lower value of average curvature indicates smooth and gradual bends, representing the sample as more flexible and shape-adaptive in nature when a soft fruit such as a tomato interacts with it.

However, a higher average curvature value indicates sharper bends, which may result in localized pressure points, which may reduce the effectiveness of the auxetic structure for uniform shape conformability over irregular shapes. The interplay between curvature and shape adaptability plays a key role in evaluating the performance of various auxetic structure samples.

After performing the experimental analysis of the various samples of the auxetic structures during the interaction with the tomato, the curvature values are investigated for the four samples, and based on the average curvature values, Aux($0^{\circ}$), with an average curvature of 1.013, exhibits the best capability for conforming the shape of the soft tomato sample. The lower value of the curvature of this sample indicates the contact is smoother, with gradual bends and uniformity, thereby reducing the possibility of high-pressure points. This makes it an ideal candidate for the gentle grasp, which can minimize the pressure distribution.

Aux($30^{\circ}$) is found with a marginally higher value of 1.034 for the average curvature, which can also perform well for good shape conformability during grasping application. However, it may not smoothly adapt the shape in comparison to Aux($0^{\circ}$). On the other hand, Aux($45^{\circ}$) is found with an average curvature value of 1.36, which is the highest among all candidates, indicating it has sharper bends and may result in uneven contact, which may have repercussions of having high pressure on the delicate surface of the fruit.

\begin{figure*}[!htbp]
    \centering
    \includegraphics[width=0.8\textwidth]{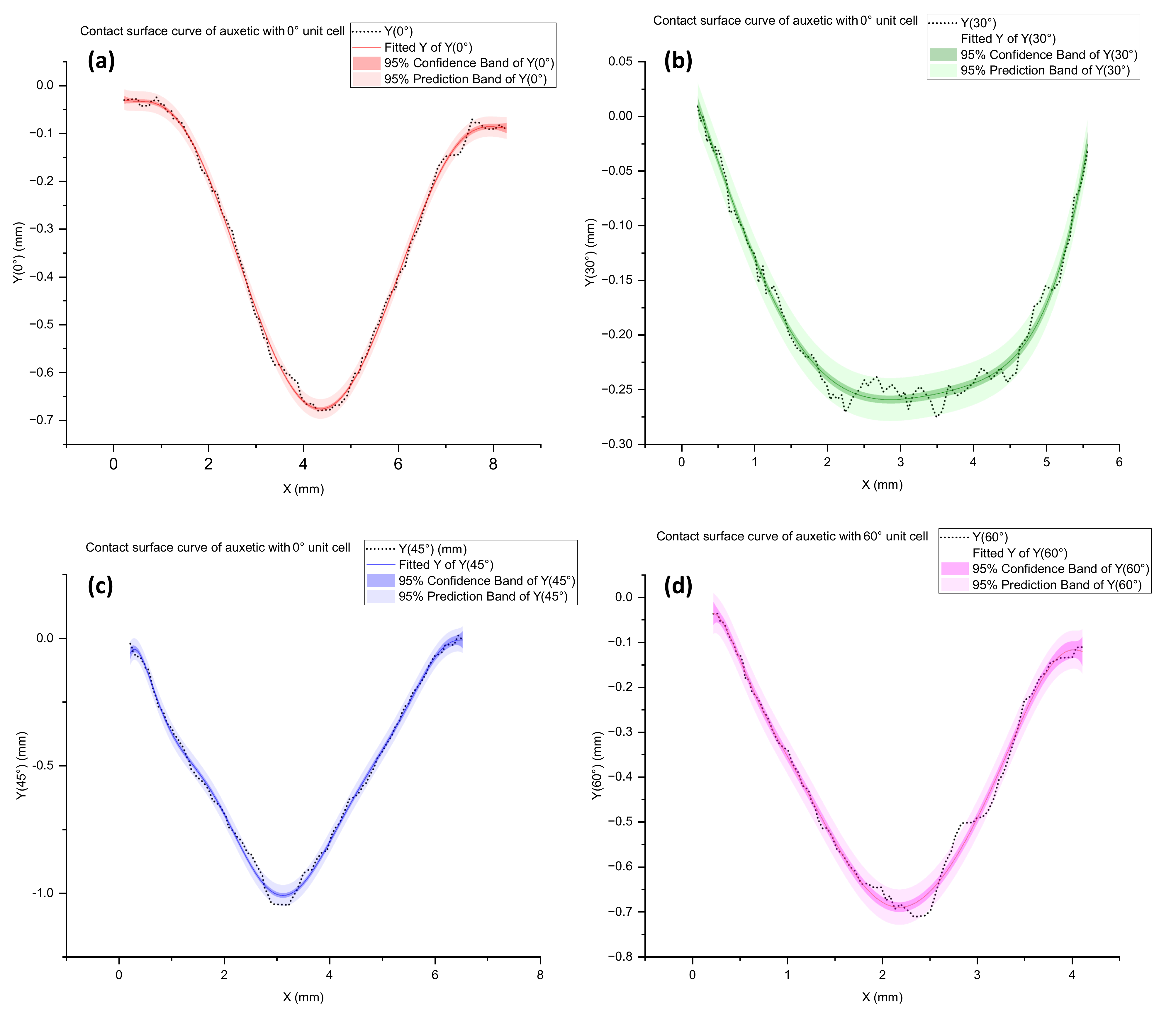}
    \caption{Plots of the deformed curves of the contact surface for the four auxetic structure samples: (a) Aux($0^{\circ}$), (b) Aux($30^{\circ}$), (c) Aux($45^{\circ}$), and (d) Aux($60^{\circ}$). These curves illustrate the relationship between displacement in the y-direction and the corresponding position along the x-axis under the effect of applied load during the interaction with the tomato.}
    \label{fig:contact_surfaces_plots}
\end{figure*}

\begin{figure*}[!htbp]
    \centering
    \includegraphics[width=0.8\textwidth]{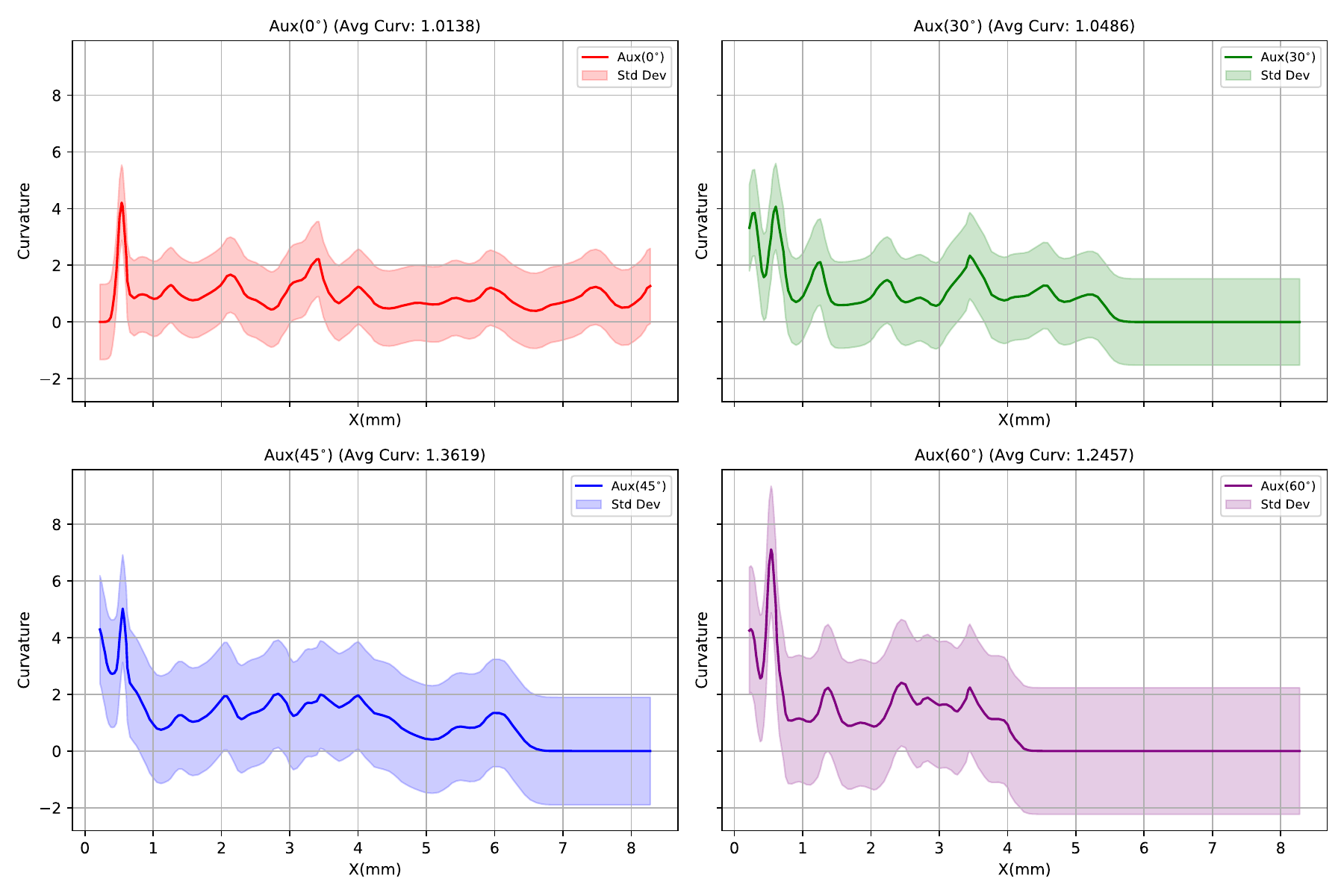}
    \caption{Figure shows the variation of the curvature of the surface of the auxetic structure samples Aux($0^{\circ}$), Aux($30^{\circ}$), Aux($45^{\circ}$) and Aux($60^{\circ}$), respectively w.r.t. x direction, obtained through experiments in 2D  while the tomato interacts with each of the samples during their individual testing.}
    \label{fig:curvature_variation_plot}
\end{figure*}

An average curvature value of 1.267 was shown by Aux($60^{\circ}$), which lies somewhat in between Aux($30^{\circ}$) and Aux($45^{\circ}$), which shows better flexibility but it has less adaptability in comparison to Aux($0^{\circ}$) and Aux($30^{\circ}$). So, it is found quantitatively that Aux($0^{\circ}$) is the best candidate among the other samples from the point of view of high shape conformability. The variation of the contact curves for samples Aux($0^{\circ}$) to Aux($60^{\circ}$) can be seen in the figure \ref{fig:contact_surfaces_plots} respectively.While the variation of curvature for these samples with respect to the x-direction can be visualized in the figure \ref{fig:curvature_variation_plot}.
\subsection{Analysis of contact forces}
When grasping a soft fruit like a tomato using a gripper, the variation in grasping force should be carefully controlled to ensure a secure hold without causing damage. Initially, the gripper should apply minimal force to gently contact the tomato, allowing it to detect the fruit's surface without deformation. The force should then be gradually increased to conform to the shape of the tomato, ensuring a firm but gentle grip. Throughout the process, force sensors can be used to monitor and adaptively adjust the applied force, maintaining a balance between securing the tomato and preventing any squashing. Once the fruit is securely grasped, the force should be kept steady to hold the tomato in place without causing compression or bruising. During release, the force should gradually be reduced to avoid sudden drops or damage.

In addition, implementing a maximum force threshold is essential to prevent the gripper from exerting excessive force that could harm the tomato. By managing the grasping force in this manner, the gripper can handle the tomato with the necessary delicacy and security. To ensure the delicate handling of tomato fruits, the experiment was performed on four different types of auxetic structure samples to evaluate their performance based on standard deviation and mean grasping force values. The variation in contact force was captured using an FSR sensor mounted on the contact surface of the auxetic structure, as shown in figure \ref{fig:Experimental_setup and schematic}.

\begin{figure*}[!htbp]
    \centering
    \includegraphics[width=0.8\textwidth]{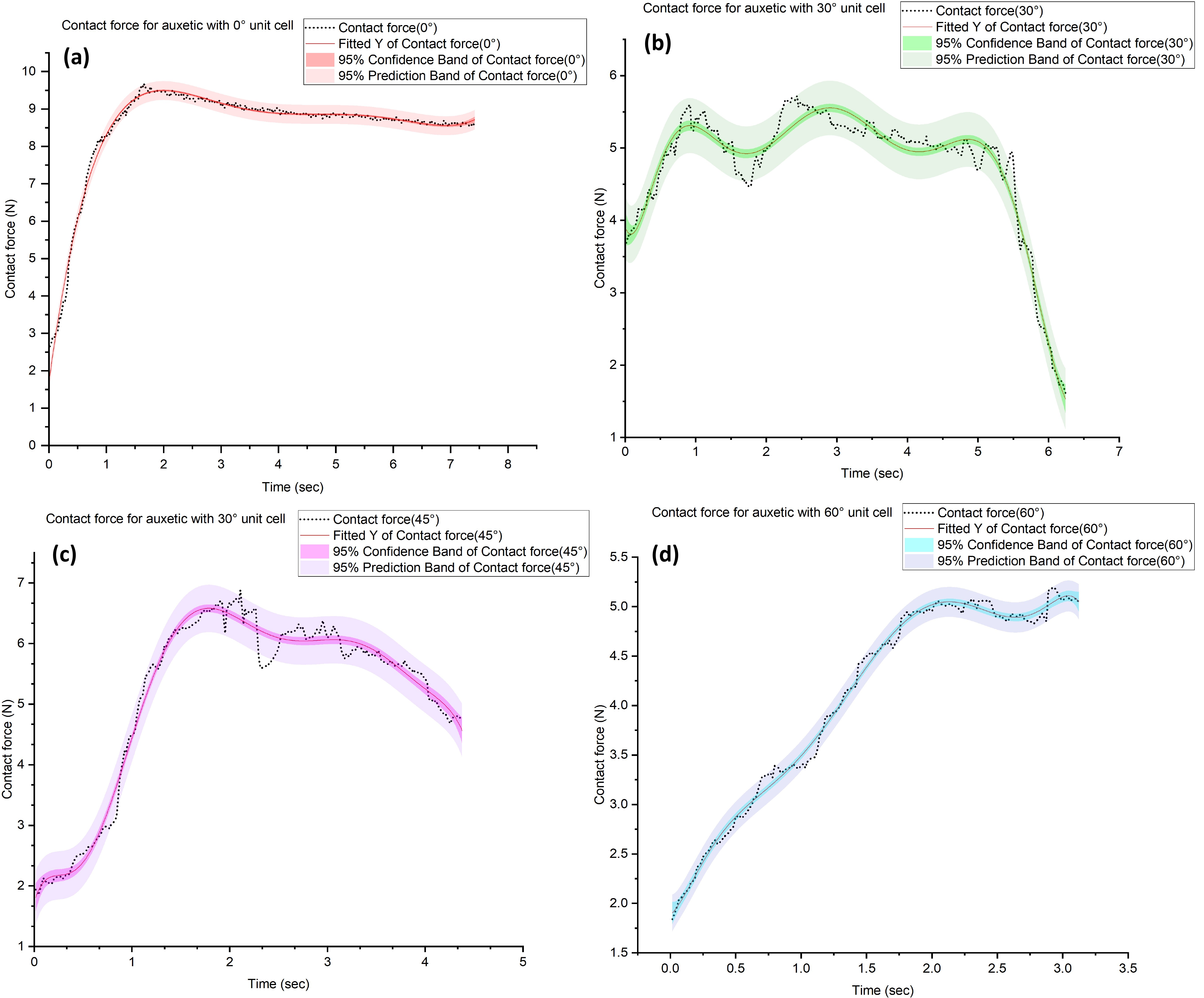}
    \caption{Plots showing the variation of contact force with respect to time for four auxetic structure samples during their interaction with a tomato under experimental conditions: (a) Aux($0^{\circ}$), (b) Aux($30^{\circ}$), (c) Aux($45^{\circ}$), and (d) Aux($60^{\circ}$). Each plot illustrates the time-dependent response during controlled testing of individual samples.}
    \label{fig:contact_forces_plots}
\end{figure*}

Here, we have used two performance parameters, which are standard deviation and mean grasping contact force, to evaluate the different auxetic structures. A low standard deviation indicates a more consistent contact force, which will be beneficial for preventing damage to the tomato fruit's exterior. On the other hand, a higher mean contact force is essential for securely holding the tomato fruit without crushing it. Based on the experimental observations, the auxetic structure Aux($0^{\circ}$) has a standard deviation of $1.31$ and mean contact force of $8.51$, which demonstrated a good balance between consistency and strength ($F_{\text{max}} = 9.59\,\text{N}$ at $t = 1.6\,\text{s}$) as depicted in Figure~\ref{fig:contact_forces_plots} (a). While the sample Aux($0^{\circ}$) ($SD: 1.31$, $mean: 8.51$) gives the most balanced combination of consistency and strength, the other auxetic structures have their own pros and cons. Auxetic structure Aux($30^{\circ}$) ($SD: 0.88$, $mean: 4.78$) has the lowest value of standard deviation, providing the most consistent contact force ($F_{\text{max}} = 5.72\,\text{N}$ at $t = 2.4\,\text{s}$), but it lacks overall strength in terms of contact force, the variation of which is shown in Figure~\ref{fig:contact_forces_plots} (b). Auxetic structure Aux($45^{\circ}$) ($SD: 1.45$, $mean: 5.16$) can offer a larger contact force ($F_{\text{max}} = 6.32\,\text{N}$ at $t = 2.7\,\text{s}$) than auxetic structure Aux($30^{\circ}$), but it has a higher value of standard deviation, indicating less consistency in the force value, as shown in Figure~\ref{fig:contact_forces_plots} (c). Meanwhile, the auxetic structure Aux($60^{\circ}$) ($SD: 1.00$, $mean: 4.09$) has a relatively low value of standard deviation and the lowest value of overall contact force ($F_{\text{max}} = 5.19\,\text{N}$ at $t = 2.9\,\text{s}$). The behavior of its variation is depicted in Figure~\ref{fig:contact_forces_plots} (d).
\noindent
Therefore, based on the criteria we have chosen for the consistency and strength in contact force, auxetic structure Aux($0^{\circ}$) will be the preferable choice for the gripper for securely grasping the tomato. The performance of this structure suggests that six such auxetic structures will be well-suited for the delicate task of tomato grasping without causing bruising or any kind of other damage.
\subsection{Finite Element Analysis of Auxetic Structures} \label{sec:FEA Analysis}
\begin{figure*}[h]
    \centering
    \includegraphics[width=0.8\textwidth]{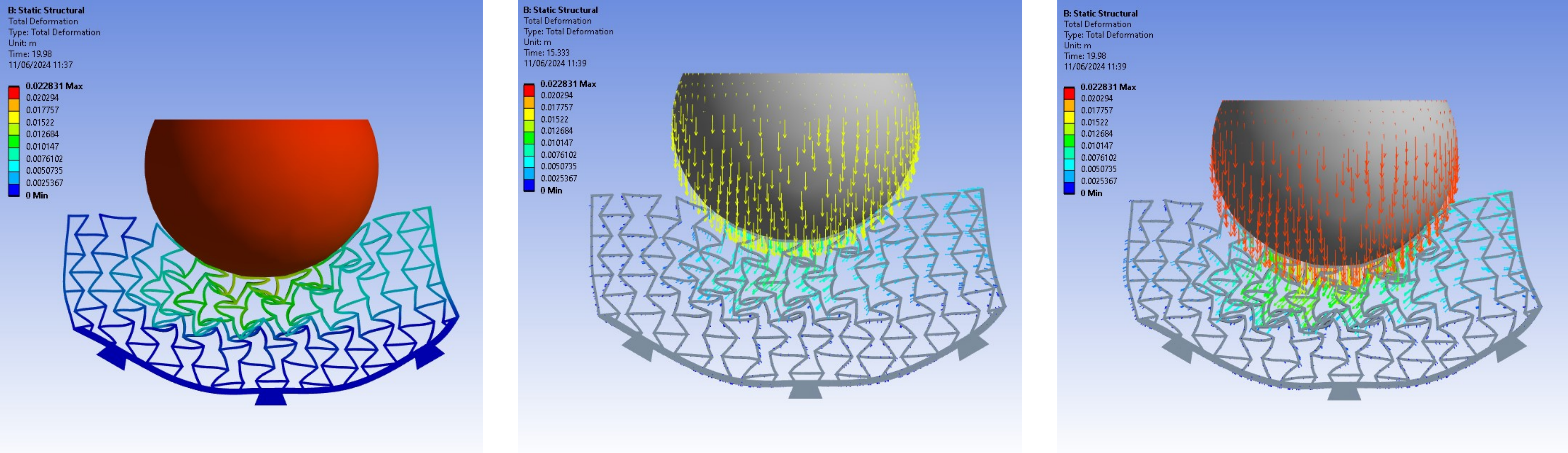}
    \caption{Figure shows the finite element analysis of auxetic structure sample Aux($0^{\circ}$) in ANSYS.}
    \label{fig:FEA_Aux_full}
\end{figure*}

The static structural analysis has been performed using the finite element analysis (FEA) package Ansys 2023 R2, as shown in Figure.\ref{fig:FEA_Aux_full}, to simulate the gripping behaviour of different auxetic samples, showing similar results as DIC and experiments. The modelling of samples is carried out in SolidWorks and imported to Ansys workbench with the material properties assigned to the simulation listed in Table. Analysis accounting for large deflections was conducted to capture nonlinear effects arising from curvature and lattice topologies. Small load increments were employed to ensure accurate resolution of lattice bending behaviour and convergence in the nonlinear analysis. The substeps are defined by a minimum of 20 to a maximum of 100 in the subspace solver, which is preferable for dealing with complex geometries. The force convergence criterion has been observed to achieve the actual forces and displacements in the system, and the number of equilibrium iterations for each subsequent substeps has been closely measured in the given time intervals. 
%%%Big figures of fea
\begin{figure*}[h]
    \centering
    \includegraphics[width=0.8\textwidth]{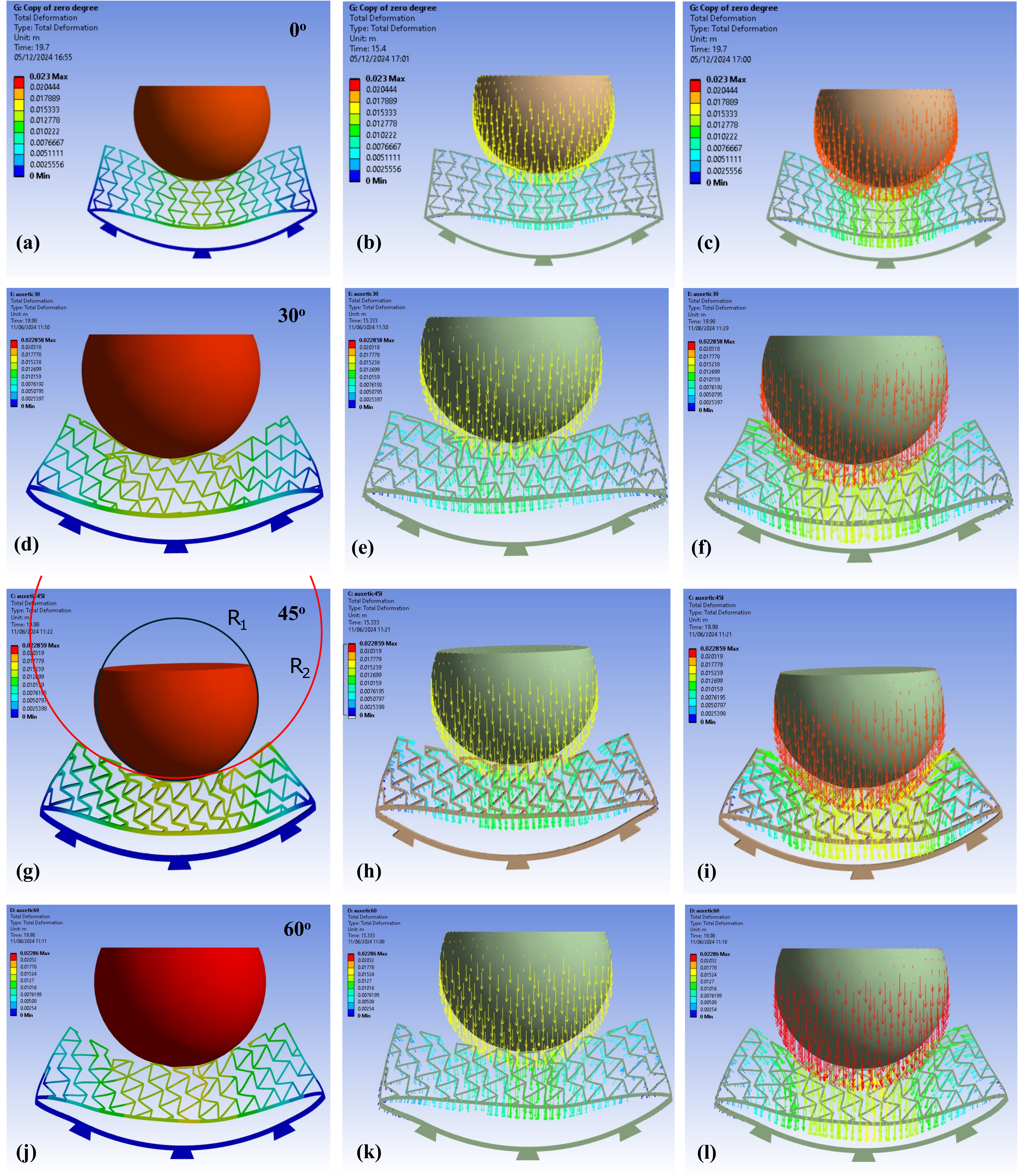}
    \caption{Figures shows deformation fields of the finite element analysis of auxetic structure sample (a-c) Aux($0^{\circ})$, (d-f) Aux($30^{\circ})$, (g-i) Aux($60^{\circ})$,
    (j-l) Aux($60^{\circ}$) in Ansys 2023 R2. The vector arrows represent the displacement field at each node or element centroid represents deformation contours overlaid with vector arrows that indicate the magnitude and direction of displacement.}
    \label{fig:curvature_plot}
\end{figure*}

\subsection{Analysis of reaction forces}
The proposed gripper combines a rigid link mechanism and soft, flexible auxetic structures. The gripper consists of six fingers, each attached to an auxetic structure, forming a cage-like structure to grip and hold the tomato firmly with a gentle grasp. The gripper finger mechanism is driven by a scotch yoke mechanism using a servo motor. Each soft auxetic structure is connected with the rigid link finger through three connecting ports, through which the forces are transmitted to it by the mechanism while grasping the tomato fruit. To determine the power required to grasp the tomato and successfully hold it in place, the grasping force and the reaction forces at the rigid fingers at the connecting ports must be determined. Also, a relationship between input torque and these forces can be developed to know the amount of torque required to provide the required grasping force for successful gripping.

To determine the reaction forces at three ports, an experimental setup is developed as shown in Fig.\ref{fig:strain_gauge_system} where on a rigid 3D-printed structure fabricated using PLA (polylactic acid) filament, three cantilever beams are fixed on the structure body and each of these beams are mounted with a single strain gauge which are connected in a quarter bridge fashion through a strain measurement device. These three beams are connected with an auxetic structure using three 3D printed connectors using white PLA filament. 

These three connections will act like three connecting ports with the rigid links of the fingers to measure the grasping force or the contact force between the auxetic structure and the tomato while it interacts with an FSR force sensor strip, which will be a type of force sensitive resistor (RP-L-110 FSR Force Sensitive Resistor Pressure Sensor). The analog voltage of sensor as output is calibrated to represent load values in Newton. Each FSR strip connects to a voltage divider circuit interfaced with ATmega328P microcontroller-based circuitry. Data from these three sensors is acquired using PLX-DAQ software. This software is a Parallax microcontroller data acquisition add-on tool that enables direct data logging from the microcontroller to Microsoft Excel.

In order to make an interaction between the tomato and auxetic structure, a linear servo motor (L16-P miniature linear actuator by Actuonix company with feedback, having stroke length 140 mm, reduction ratio 63:1, and operating voltage 12 volts) is actuated, which is connected with a 3D printed conical tomato holder that holds the tomato. When the tomato interacts with the auxetic structure, the contact force can be read by the FSR sensor, the values of which can be directly captured via PLX DAQ on the computer. Simultaneously, the reaction force values at the connecting ports can be read by the three strain gauges mounted on the respective cantilever beam, which can be stored as microstrains in the computer storage through the data acquisition system and can be transformed in terms of force values as discussed in detail in appendix \ref{appendix:strain_measurement_process}.

\begin{figure*}[!htbp]
    \centering
    % First subfigure
    \begin{subfigure}[b]{0.40\textwidth}
        \includegraphics[width=\textwidth]{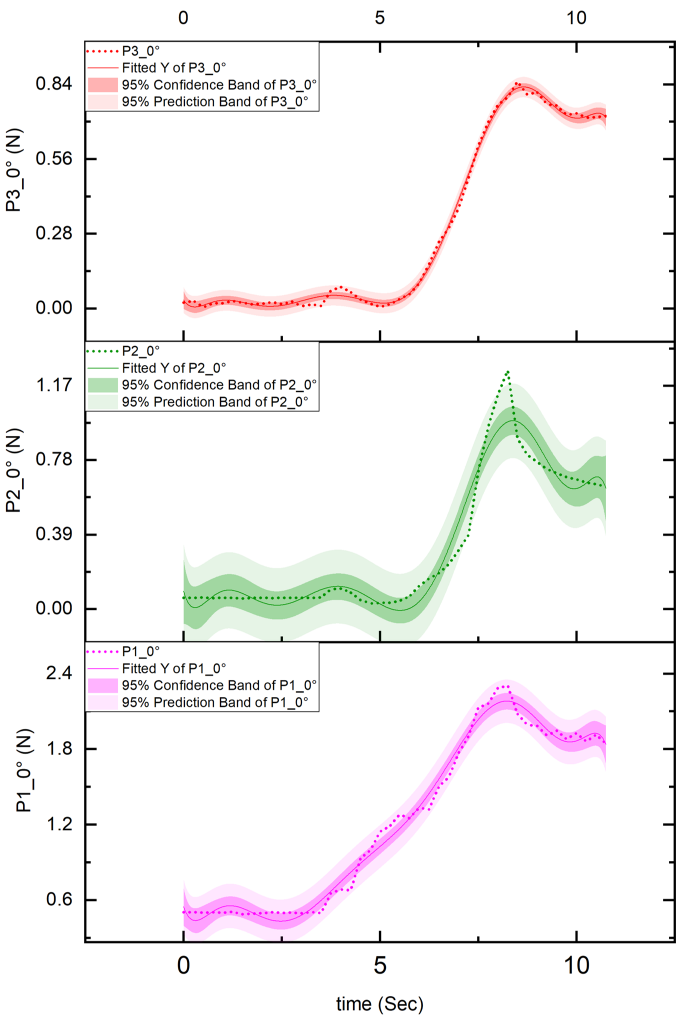}
        \begin{picture}(0,0)
            \put(-10,160){(a)}
        \end{picture}
        \caption{$0^{\circ}$}
        \label{fig:strain_gauge_forces_0deg}
    \end{subfigure}
    \hspace*{1em} % Reduced horizontal spacing
    % Second subfigure
    \begin{subfigure}[b]{0.40\textwidth}
      \includegraphics[width=\textwidth]{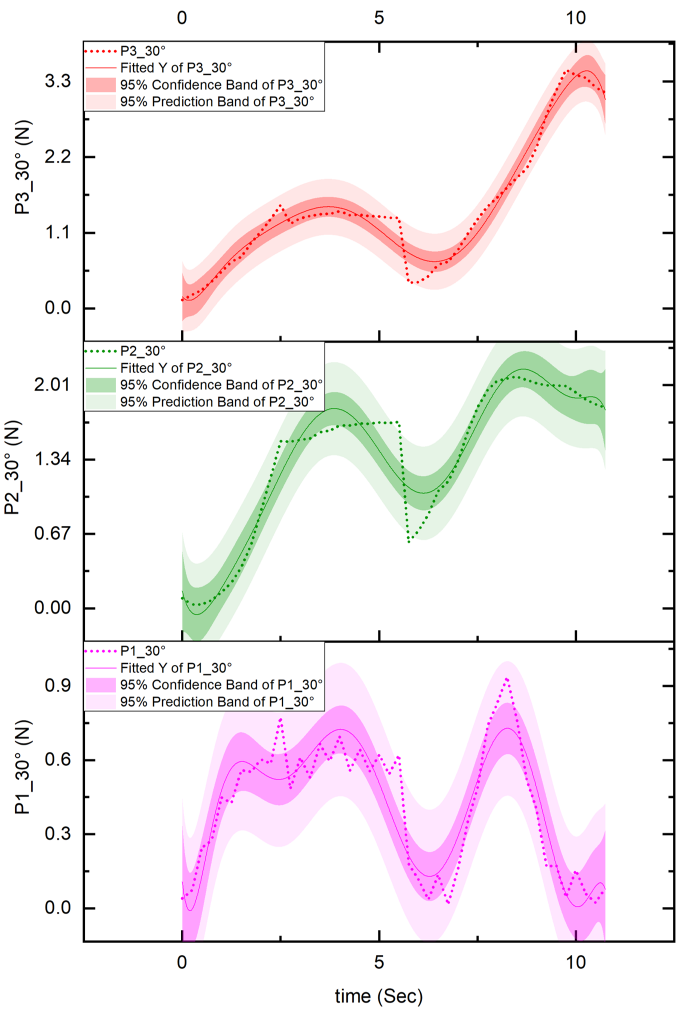}
        \begin{picture}(0,0)
            \put(-10,160){(b)}
        \end{picture}
        \caption{$30^{\circ}$}
        \label{fig:strain_gauge_forces_30deg}
    \end{subfigure}
    
    \vspace{-1em}
    % Third subfigure
    \begin{subfigure}[b]{0.40\textwidth}
        \includegraphics[width=\textwidth]{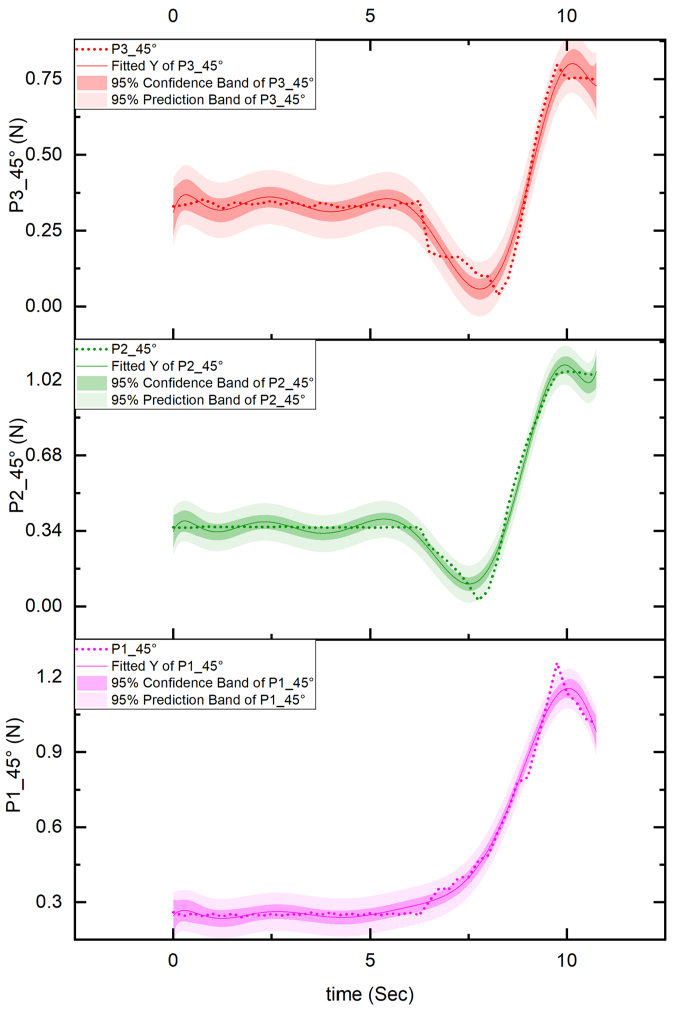}
        \begin{picture}(0,0)
            \put(-10,160){(c)}
        \end{picture}
        \caption{$45^{\circ}$}
        \label{fig:strain_gauge_forces_45deg}
    \end{subfigure}
    \hspace*{1em} % Reduced horizontal spacing
    % Fourth subfigure
    \begin{subfigure}[b]{0.40\textwidth}
        \includegraphics[width=\textwidth]{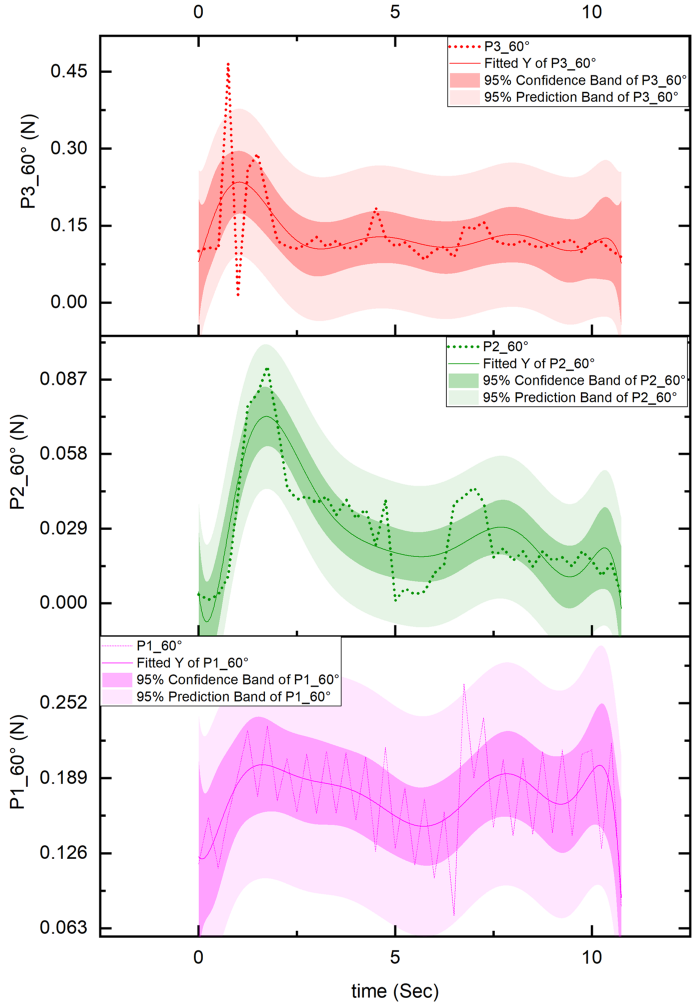}
        \begin{picture}(0,0)
            \put(-10,160){(d)}
        \end{picture}
        \caption{$60^{\circ}$}
        \label{fig:strain_gauge_forces_60deg}
    \end{subfigure}
  
    \caption{Plots showing the variation of reaction forces at the three connecting ports of the auxetic structures with varying unit cell inclinations: 
    (a) $0^{\circ}$, (b) $30^{\circ}$, (c) $45^{\circ}$, and (d) $60^{\circ}$. The results are derived from experimental analysis and demonstrate the differences in force distributions for each auxetic configuration.}
    \label{fig:Reaction forces}
\end{figure*}

To experimentally obtain the values of three reaction forces that the auxetic structure is encountering via three connecting ports of the rigid fingers of the mechanism, we experimented on the developed experimental setup upon which the auxetic structure interacts with the three cantilever beams mounted with a strain gauge on each of them. The values of these reaction forces for different auxetic structures with different unit cell inclinations can be obtained, which forms the basis for the reaction forces that rigid link fingers of the gripper mechanism encounter during the interaction of the tomato fruit while grasping it in the actual gripper mechanism. From these reactions, an estimate of the required motor torque to maintain the desired grasping force by the mechanism can be determined, which will form the basis for performing the grasping experiments on the tomato fruits in the actual case scenario of the gripper operation while considering the weight and frictional effects.
The results in figure \ref{fig:Reaction forces} show the variation of three reaction forces $P_1$, $P_2$, and $P_3$ with respect to the time for various samples having different unit cell inclinations with respect to the vertical axis. The variation of the contact forces for these auxetic structures can be depicted in figure \ref{fig:contact_forces_plots}. The peak values of the reaction forces and the contact forces reached for various auxetic structure samples during the interaction with the tomato sample is obtained experimentally, which can be seen in table \ref{table:Reaction_and_contact_forces} which is based upon figure \ref{fig:strain_gauge_forces_0deg},\ref{fig:strain_gauge_forces_30deg},\ref{fig:strain_gauge_forces_45deg} and \ref{fig:strain_gauge_forces_60deg} respectively.
\begin{table*}[t]
\small\sf\centering
\caption{Peak values of reaction and contact forces for various auxetic structure samples.\label{table:Reaction_and_contact_forces}}
\begin{tabular}{lcccc}
\toprule
Peak force value (N) & Aux($0^\circ$) & Aux($30^\circ$) & Aux($45^\circ$) & Aux($60^\circ$) \\
\midrule
$P_{1}$ & 0.86 & 0.93 & 1.25 & 0.26 \\ 
$P_{2}$ & 1.25 & 2.08 & 1.05 & 0.09 \\
$P_{3}$ & 2.35 & 3.46 & 0.79 & 0.46 \\
$P_{c}$ & 9.65 & 5.60 & 4.51 & 3.41 \\
\bottomrule
\end{tabular}
\end{table*}
\begin{table*}[t]
\small\sf\centering
\caption{Statistical analysis of forces.\label{tab:Statistical_Forces}}
\setlength{\tabcolsep}{4pt} % adjust padding if needed
\begin{tabular}{@{}lccccc@{}}
\toprule
Force & Number of Points (N) & Degree of Freedom (DF) & Residual Sum of Squares (RSS) & R-square (R\(^2\)) & Adjusted R-square (Adj R\(^2\)) \\
\midrule
$P_{1}(0^{\circ})$  & 44 & 34 & 0.20861 & 0.98901 & 0.9861 \\
$P_{2}(0^{\circ})$  & 44 & 34 & 0.26973 & 0.95084 & 0.93782 \\
$P_{3}(0^{\circ})$  & 44 & 34 & 0.01024 & 0.99784 & 0.99727 \\
$P_{1}(30^{\circ})$ & 44 & 34 & 0.52171 & 0.82380 & 0.77716 \\
$P_{2}(30^{\circ})$ & 44 & 34 & 1.25887 & 0.93373 & 0.91619 \\
$P_{3}(30^{\circ})$ & 44 & 34 & 1.18413 & 0.96866 & 0.95960 \\
$P_{1}(45^{\circ})$ & 44 & 34 & 0.04107 & 0.99031 & 0.98775 \\
$P_{2}(45^{\circ})$ & 44 & 34 & 0.04982 & 0.98487 & 0.98087 \\
$P_{3}(45^{\circ})$ & 44 & 34 & 0.05660 & 0.96263 & 0.95273 \\
$P_{1}(60^{\circ})$ & 44 & 34 & 0.06151 & 0.26652 & 0.07236 \\
$P_{2}(60^{\circ})$ & 44 & 34 & 0.00542 & 0.73863 & 0.66944 \\
$P_{3}(60^{\circ})$ & 44 & 34 & 0.13837 & 0.29330 & 0.10623 \\
\bottomrule
\end{tabular}
\end{table*}
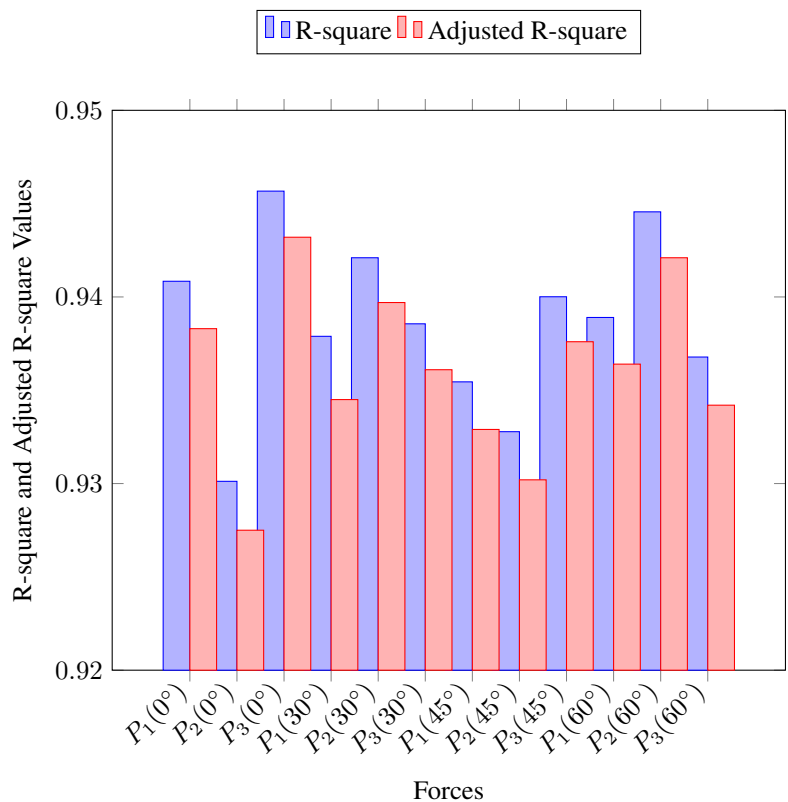
\begin{figure*}[!htbp]
\centering
\begin{tikzpicture}
\begin{axis}[
    width=0.6\textwidth,
    ybar=0pt,
    bar width=10pt,
    enlarge x limits=0.15, % Slight spacing between groups
    xlabel={Forces},
    ylabel={R-square and Adjusted R-square Values},
    symbolic x coords={$P_1(0^{\circ})$, $P_2(0^{\circ})$, $P_3(0^{\circ})$, $P_1(30^{\circ})$, $P_2(30^{\circ})$, $P_3(30^{\circ})$, $P_1(45^{\circ})$, $P_2(45^{\circ})$, $P_3(45^{\circ})$, $P_1(60^{\circ})$, $P_2(60^{\circ})$, $P_3(60^{\circ})$},
    xtick=data,
    ymin=0.92, ymax=0.95,
    xticklabel style={rotate=45, anchor=east}, % Rotated x-tick labels
    legend style={at={(0.5,1.1)}, anchor=south,legend columns=-1}, % Legend above
    ytick={0.92, 0.93, 0.94, 0.95}, % Explicit y-axis ticks
    ]
% Grouped bars for R-square and Adjusted R-square
\addplot coordinates {
    ($P_1(0^{\circ})$,0.94084) ($P_2(0^{\circ})$,0.93012) ($P_3(0^{\circ})$,0.94567)
    ($P_1(30^{\circ})$,0.93789) ($P_2(30^{\circ})$,0.94210) ($P_3(30^{\circ})$,0.93856)
    ($P_1(45^{\circ})$,0.93545) ($P_2(45^{\circ})$,0.93278) ($P_3(45^{\circ})$,0.94001)
    ($P_1(60^{\circ})$,0.93890) ($P_2(60^{\circ})$,0.94456) ($P_3(60^{\circ})$,0.93678)
};
\addplot coordinates {
    ($P_1(0^{\circ})$,0.93830) ($P_2(0^{\circ})$,0.92750) ($P_3(0^{\circ})$,0.94320)
    ($P_1(30^{\circ})$,0.93450) ($P_2(30^{\circ})$,0.93970) ($P_3(30^{\circ})$,0.93610)
    ($P_1(45^{\circ})$,0.93290) ($P_2(45^{\circ})$,0.93020) ($P_3(45^{\circ})$,0.93760)
    ($P_1(60^{\circ})$,0.93640) ($P_2(60^{\circ})$,0.94210) ($P_3(60^{\circ})$,0.93420)
};

\legend{R-square, Adjusted R-square}
\end{axis}
\end{tikzpicture}
\caption{Comparison of R-square and Adjusted R-square values for different forces and angles. Bars are grouped for easy comparison.}
\label{fig:Auxetic_bar_graph}
\end{figure*}
In this study, experimental force data for 12 different cases of reaction forces ($P_1(0^{\circ})$ to $P_3(60^{\circ})$) were plotted using 2D graphs in OriginLab data analysis and graphing software, each with a 95\% confidence band and a 95\% prediction band. These statistical bands provide important insight into the range of variability in the measurements and the confidence in the regression line fitted to the data. To quantitatively assess the quality of the fit, the R-square ($R^2$) and Adjusted R-square ($Adj R^2$) values were calculated for each force case. 

For instance, for force $P_1(0^{\circ})$, with an $R^2$ value of 0.94084 and an $Adj R^2$ value of 0.93830, it is evident that over 94\% of the variance in the data is explained by the fitted regression line, suggesting a strong correlation between the experimental measurements and the predicted values. Additionally, the minimal difference between $R^2$ and $Adj R^2$ (approximately 0.0025) indicates that the model is not overfitted, implying that the predictors used in the regression line are well-chosen and do not inflate the model's performance. This pattern holds across most of the forces, with all $R^2$ values exceeding 0.90, confirming a high degree of accuracy in the data fit.

The residual sum of squares ($RSS$) for $P_1(0^{\circ})$, reported as 0.26872, represents the total deviation of the observed values from the fitted regression line. A lower $RSS$ value across the forces, particularly in the case of  $P_1(0^{\circ})$, further strengthens the reliability of the model fit. Moreover, with a degree of freedom of 32 for  $P_1(0^{\circ})$, the model accounts for sufficient experimental variability, ensuring that the conclusions drawn from the data are robust.

The inclusion of 95\% confidence and prediction bands further quantifies the uncertainty in the regression line and the likely range for future observations. The confidence band indicates the precision of the estimated regression line, while the prediction band illustrates the expected spread of future data points. In general, the high $R^2$ and $Adj R^2$ values, low $RSS$ values, and confidence intervals collectively indicate that the experimental force data are well represented by the fitted regression lines, and the prediction bands provide a reliable indication of how the system might behave under similar conditions in the future.

%%%Bar graph discussion
The bar graph illustrates the values of R-square ($R^2$) and adjusted R-square ($Adj R^2$) for the 12 experimental reaction forces ($P_1(0^{\circ}$) to $P_3(60^{\circ})$), providing a quantitative analysis of how well the regression lines fit the observed experimental data. For example, the $R^2$ value for force $P_1(0^{\circ})$ is 0.94084, indicating that 94\% of the variance in the data is explained by the fitted regression line, with a corresponding $Adj R^2$ value of 0.9383, suggesting minimal overfitting. Similar trends are observed for other forces, with $R^2$ values consistently above 0.90, demonstrating strong correlations between the experimental force data and the regression models. 

\indent Notably, force $P_3(0^{\circ})$ exhibits one of the highest $R^2$ values at 0.9532, reflecting a particularly strong data fit. The minimal difference between $R^2$ and $Adj R^2$ across all forces, typically less than 0.003, confirms the robustness of the models without unnecessary complexity. This quantitative analysis confirms that the experimental data closely follow the predicted trends, with low residuals and high confidence in the regression lines, underscoring the reliability and accuracy of the experimental force measurements. The bar graph effectively highlights these trends, providing a clear visual comparison of the goodness of fit across the various force conditions studied.
\subsection{Comparative Study of Curvatures of Auxetic Structures Using DIC and FEA}
The curvature assessment of auxetic structures plays a pivotal role in evaluating their capability to conform to soft, irregular surfaces such as fruits. Two different methods, namely Digital Image Correlation (DIC) and Finite Element Analysis (FEA), were employed to analyze curvature in four auxetic samples, labeled Aux($0^\circ$), Aux($30^\circ$), Aux($45^\circ$), and Aux($60^\circ$). While DIC measures curvature directly from the experimental contact profile of the auxetic structures against the fruit surface, FEA simulates the contact interaction under controlled boundary conditions. Both methods yield insights into shape conformability, though they rely on different metrics for quantifying curvature.

In the DIC approach, the local curvature $k$ of each contact profile $y = f(x)$ is computed using the formula 
\[
k = \frac{\bigl|y''\bigr|}{\bigl(1 + (y')^2 \bigr)^{3/2}},
\]
where $y' = \tfrac{dy}{dx}$ and $y'' = \tfrac{d^2y}{dx^2}$ represent the first and second derivatives of the contact curve, respectively. By examining the average curvature of each auxetic sample, a lower value of this metric indicates smoother bending and better adaptability to the tomato’s surface. Accordingly, Aux($0^\circ$) exhibited the lowest average curvature of about 1.013, highlighting its gentle contact profile. Aux($30^\circ$) was slightly higher at 1.034, showing reasonably good conformability but with marginally sharper bends than Aux($0^\circ$). Aux($45^\circ$) had the highest average curvature (1.360), suggesting the presence of more localized deformation, while Aux($60^\circ$) fell between these extremes at 1.267. Based on this experimental evidence, Aux($0^\circ$) emerged as the most desirable design for gentle and uniform pressure distribution on the delicate fruit surface.

In contrast, the FEA study introduces two radii, $R_1$ for the tomato and $R_2$ for the auxetic samples, and defines the curvature ratio 
\[
\kappa_r = \frac{R_1}{R_2}.
\]
A $\kappa_r$ value of 1 implies near-perfect conformity between the auxetic structure and the fruit’s curvature. The simulation reveals that Aux($45^\circ$) achieves the highest curvature ratio of 0.61, indicating superior overall matching. Aux($60^\circ$) follows at 0.53, then Aux($0^\circ$) at 0.51, and finally Aux($30^\circ$) at 0.48. Hence, from an FEA standpoint, Aux($45^\circ$) appears most effective at caging or ``wrapping'' the fruit, while Aux($30^\circ$) demonstrates the least conformability under these simulation conditions.

When comparing the experimental and simulation findings, a discrepancy in rankings emerges. The DIC measurements indicate that Aux($0^\circ$) delivers the smoothest contact profile and presumably minimal pressure localization, making it ideal for gentle handling. Conversely, the FEA results highlight Aux($45^\circ$) as best at conforming to the tomato’s overall curvature, suggesting strong caging potential. These differences can stem from factors such as simplified boundary conditions in the simulation, material property assumptions, and variances in friction or alignment during physical experiments. Ultimately, for delicate fruit handling where bruising is critical, Aux($0^\circ$) might be the prime candidate based on its low curvature profile observed experimentally. However, if maximizing the gripping or caging effect is the priority---such as ensuring the fruit does not slip---Aux($45^\circ$) is favored according to the FEA-derived curvature ratio. An optimal design approach would incorporate both perspectives, maintaining a balance between gentle handling and secure gripping of the fruit.

\subsection{Interaction analysis of auxetic structures with rigid link mechanism}
Since the auxetic structure is attached to each finger of the gripper's rigid link mechanism via three ports, the reaction forces will be transmitted by the compliant auxetic structures to the rigid link mechanism during the tomato grasping process. The servo motor provides the necessary torque to hold the tomato securely without losing contact. After testing various auxetic structure samples for the reaction forces they encounter, we have collected data on the three reaction forces for each structure, which vary with respect to time. Based on these reaction force data, we evaluated the necessary torque required to hold the tomato firmly during grasping and compared how these torques vary with respect to changes in the crank angle of the scotch yoke mechanism. 

\begin{figure*}[h!]
    \centering
    \includegraphics[width=0.9\textwidth]{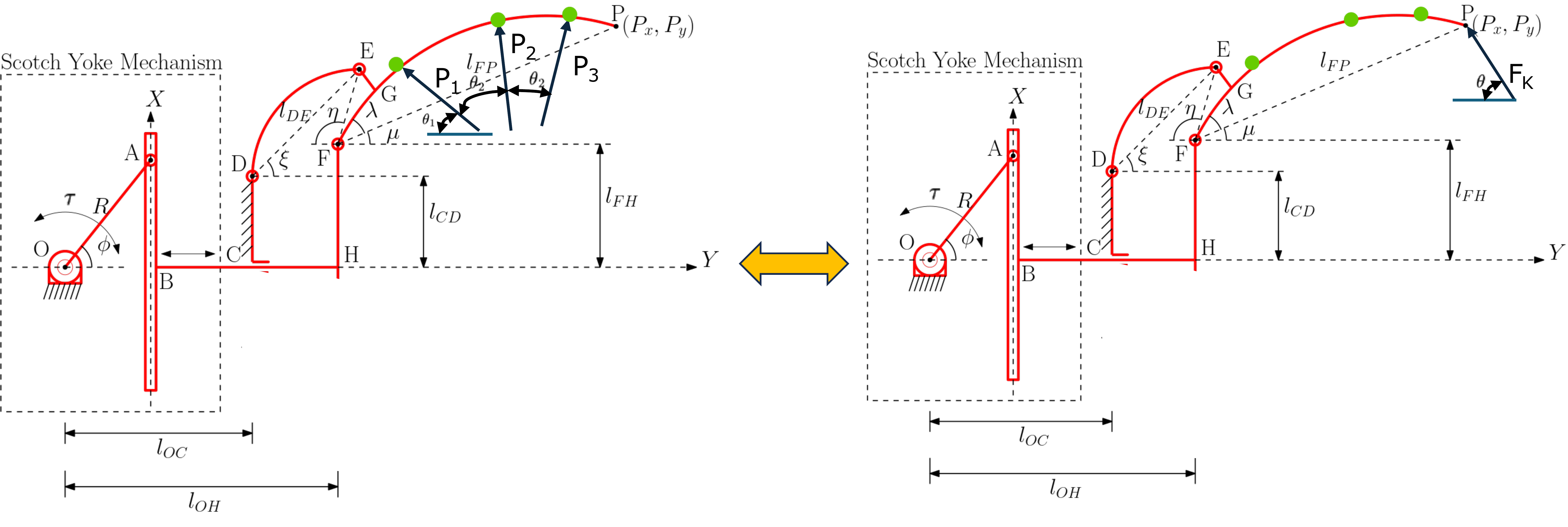}  % Adjust the width as needed
    \caption{The figure depicts the gripper rigid link mechanism interacting with reaction forces applied by auxetic structure connecting ports (the left-hand side of the figure shows the actual reaction forces acting, while right-hand side of this figure shows its equivalent static diagram with a single force on the tip) \cite{ansari2025design} while grasping the tomato along with the motor torque to provide necessary grasping force.}
    \label{fig:mechanism_force_interaction}
\end{figure*}

The force of interactions can be seen in the rigid link mechanism diagram shown in the figure \ref{fig:mechanism_force_interaction} while its geometric parameters can be depicted from the table   
\ref{tab:geometric_properties}. These reaction forces $P_1$,$P_2$ and $P_3$ are measured experimentally through three strain gauges mounted on three connecting ports of the auxetic structure (can be seen in the left of the figure \ref{fig:mechanism_force_interaction}) while in actual these reaction forces will be experienced by the rigid link mechanism as shown in the figure \ref{fig:strain_gauge_system} and the equivalence of these three forces can be written as a single point force $F_k$ acting at the tip at a point P of the rigid finger (can be seen in the right of the figure \ref{fig:mechanism_force_interaction}) as mentioned in expression 31 at an angle $\theta$ which can be obtained from expression 32 ( refer to section 5 of supplementary material)).
\begin{table}[h!]
\centering
\caption{Geometric properties of rigid link mechanism (\cite{ansari2025design})}
\label{tab:geometric_properties}%
\begin{adjustbox}{max width=\textwidth}
\begin{tabular}{lcc}
\toprule
Parameter       & Vaue \\
\midrule
$L_{EG}$        & $42.72 mm$   \\
$R$             &  $30 mm$      \\
$R$             &  $30 mm$      \\
$L_{OC}$        & $128.50 mm$  \\
$L_{DE}$        & $59.78 mm$      \\ 
$L_{FE}$        & $42.73 mm$          \\ 
$L_{FP}$        & $110.48 mm$     \\ 
$L_{CD}$        & $40.09 mm$      \\
$L_{FH}$        & $43.08 mm$      \\ 
$\phi$          & $0^{o}-90^{o}$          \\
$\xi$           & $0^{o}-90^{o}$          \\ 
$\eta$          & $\text{-}120^{o}-\text{-}90^{o}$ \\ 
$\lambda$       & $42^{o}$  \\ 
$\mu$           & $0^{o}-90^{o}$          \\
$\nu$           & $14^{o}$          \\ 
\bottomrule
\end{tabular}
\end{adjustbox}
\end{table}

Using the geometry and symmetry of the gripper finger, the angle between the normals can be found to be symmetric, which is $\theta_2=30^\circ$. Here, we have performed the static force analysis of the gripper mechanism by considering a single finger driven by a scotch yoke mechanism through a servo motor. While interacting with a tomato-like soft fruit with its soft auxetic part, the grasping contact force will be transmitted from the contact surface of the auxetic structure to the rigid finger via three connecting ports. To maintain these forces, the servo motor has to provide sufficient torque to hold the tomato in place. 

The complete free-body diagram of the gripper mechanism is shown in the figure. Since the gripper mechanism is symmetric, it is assumed here that the forces we are getting at three ports will be the maximum force that a rigid finger will encounter while interacting with the tomato under the extreme case scenario of auxetic deformation, and to maintain these maximum values of forces on each of the six fingers of the gripper, the motor has to provide the torque six times the torque required for a single finger to hold the tomato under the extreme condition.
\subsubsection{Force-torque relationship}
\begin{figure*}[!htbp]
  \centering
  \includegraphics[width=0.7\textwidth]{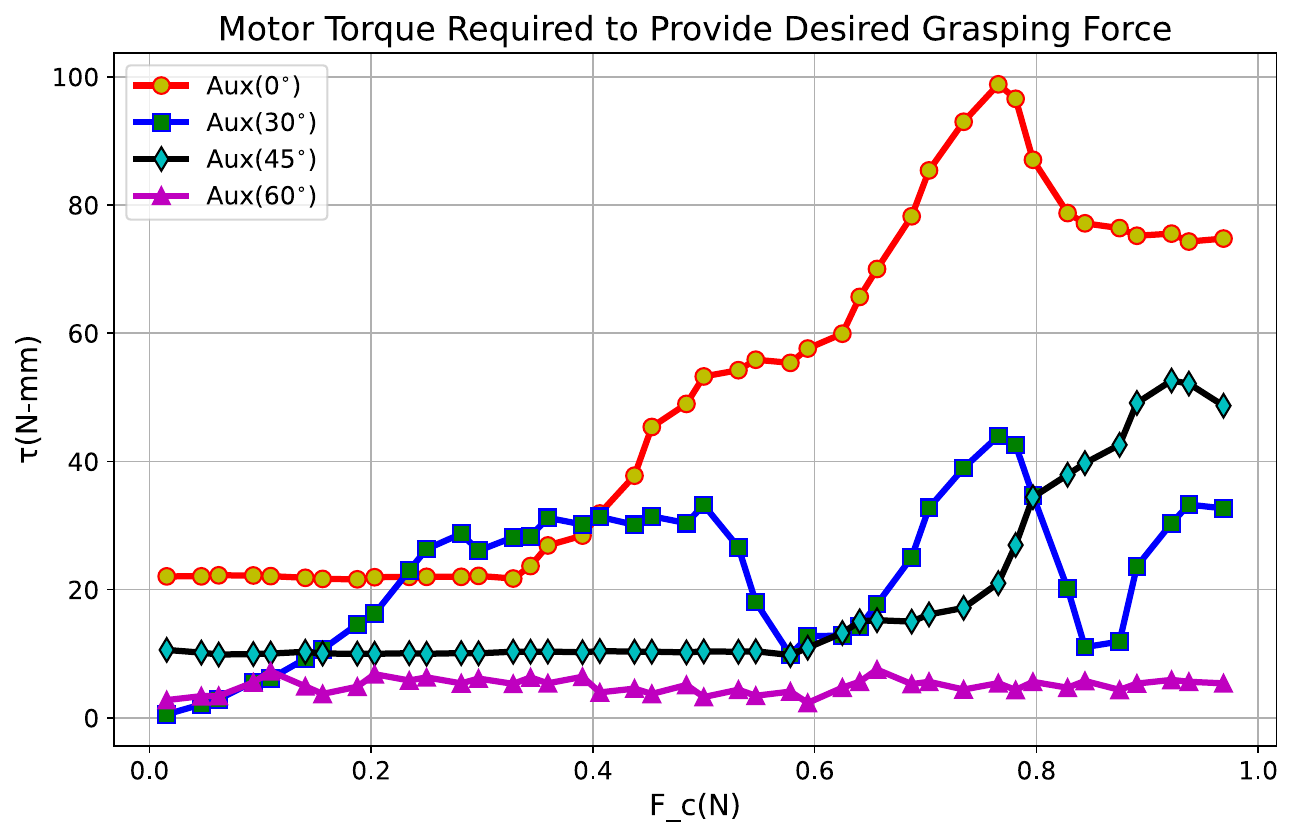}
  \caption{Shows the plot of required motor torque versus grasping force for four different auxetic structure samples. The curves represent how the motor torque varies with increasing grasping force across different auxetic samples. Each curve corresponds to a specific sample, highlighting differences in torque efficiency and force transmission. The plot provides insights into the mechanical performance of the structures during the interaction experiment with the tomato, with some samples requiring higher torque for the same grasping force, indicating differences in mechanical advantage or frictional characteristics. This comparison is crucial for optimizing motor power requirements for effective grasping.}
  \label{fig:torque_vs_grasping_force}
\end{figure*}

We have developed a relation between the reaction forces and the torque based on the static analysis of the rigid link mechanism, as can be depicted in the figure. This expression consists of various parameters, such as the equivalent reaction force, which is supposed to be acting on the fingertip, and some fixed geometric parameters, the details of which can be seen in the table.
\begin{equation}
\tau = F_k R \sin\phi \left(
    \frac{l_{FP}\,\sin(\mu-\theta)\,\cos\zeta}{\,l_{FE}\,\sin(\eta+\xi)\,}
    - \cos\theta
\right)
\end{equation}
This expression can be used to evaluate the variation in motor torque required to securely hold the target tomato based on the experimentally obtained reaction forces, as shown in Figure \ref{fig:Tomato_interaction_reaction forces}. It should be noted that this experiment was conducted using a tomato with a nominal diameter of approximately 5.3 cm under the maximum safe load that the linear actuator can provide. Based on these conditions, we obtained the reaction forces. However, the variation of these reaction forces for auxetic structures with different unit cells can be depicted in figure \ref{fig:Reaction forces}. Based upon the evaluation of $F_k$, $\theta$, and other geometric parameters, we can evaluate the variation of the motor torque for the four different auxetic structures. The variation of these torques can be seen in the figure, which is discussed quantitatively in section \ref{fig:torque_vs_grasping_force}
\subsubsection{Analysis of torque for auxetic structures with different unit cells}
Comparing the four auxetic structure samples regarding motor torque demands at different grasping forces offers valuable insights for soft tomato fruit harvesting. Each sample reveals unique torque characteristics over grasping forces ranging from 0.2 N to 0.8 N, helping assess their suitability for handling delicate produce. The Aux($60^{\circ}$) sample, for instance, exhibits a stable torque profile, with torque values decreasing from 6.73 N-mm at 0.2 N to 3.36 N-mm at 0.6 N, then slightly increasing to 5.58 N-mm at 0.8 N. This trend suggests optimal flexibility and adaptability for delicate handling with low energy consumption, which is crucial to minimizing damage to sensitive fruits. The consistently low torque levels indicate that Aux($60^{\circ}$) is well-suited for applications demanding gentle force control, making it a strong candidate for soft tomato fruit harvesting.

In contrast, Aux($45^{\circ}$) shows a steadily increasing torque demand, from 10.47 N-mm at 0.2 N to 34.8 N-mm at 0.8 N, which implies a firmer grip. This could be beneficial for sturdier tomato fruits but might lead to excessive pressure on softer ones, especially at higher grasping forces. Hence, Aux($45^{\circ}$) could be considered for slightly tougher produce, where a firmer grip is less likely to cause harm. The Aux($30^{\circ}$) sample demonstrates significant torque variability, beginning at 17.17 N-mm at 0.2 N, peaking at 30.75 N-mm at 0.4 N, and reaching 35.7 N-mm at 0.8 N. This variability may result in uneven pressure on the fruit, making it less ideal for handling soft produce, where consistency is crucial to avoid bruising.
Lastly, Aux($0^{\circ}$) requires the highest torque values, starting at 22.62 N-mm at 0.2 N and increasing sharply to 87.2 N-mm at 0.8 N, indicating a design intended for robust gripping applications. However, this rigidity would likely be unsuitable for delicate fruits, as it could lead to significant surface pressure compared to the other samples.

Overall, Aux($60^{\circ}$) appears most suitable for soft tomato fruit harvesting, offering low and stable torque essential for gentle handling. Conversely, Aux($45^{\circ}$), Aux($30^{\circ}$), and Aux($0^{\circ}$) might be better suited for firmer fruits or less delicate harvesting applications, where higher or variable gripping forces are acceptable. This analysis highlights the importance of choosing materials with controlled force characteristics to protect soft tomato fruit integrity during harvest.
% \FloatBarrier
\section{Conclusion}
This study highlights the mechanical advantages of auxetic structures with varying unit cell inclinations for soft fruit handling applications. Among the designs tested, the Aux($45^{\circ}$) sample demonstrated superior flexibility and deformation capacity, with the highest strain (0.108) and displacement (12.5 mm) values. The Aux($0^{\circ}$) structure excelled in shape conformability, achieving an average curvature of 1.013, and exhibited balanced contact force behavior (mean 8.51 N, peak 9.59 N, standard deviation 1.31 N), making it particularly suitable for securely grasping delicate tomatoes with minimal pressure-induced damage.

The Aux($60^{\circ}$) sample exhibited a stable, adaptive torque profile, suggesting its potential for energy-efficient and controlled grasping under variable loading conditions. Overall, each auxetic configuration demonstrated unique strengths, supporting the potential for customized gripper designs based on specific agricultural requirements.

While early-stage modeling employed small deformation assumptions for analytical simplicity, nonlinear finite element simulations (Section~\ref{sec:FEA Analysis}) were incorporated to capture large-deformation behaviors more accurately. Future work will further refine dynamic models to better represent real-world handling conditions.

Additionally, the current study did not address eigenvalue buckling analysis, which remains critical for evaluating structural stability in highly deformable geometries. Future investigations will integrate eigenbuckling simulations, following frameworks such as Adhikari et al.
 (\cite{adhikari2021eigenbuckling}), to enhance gripper reliability and safety.
\begin{acks}
This research was supported by the Department of Science and Technology, Government of India, under project number DST/ME/2020009. The authors thank Professor P. Venkitanarayanan from the High-Speed Lab in the Mechanical Engineering Department at IIT Kanpur, India, for providing the experimental facility for Digital Image Correlation.
\end{acks}

% \begin{thebibliography}{99}
% \bibitem[Kopka and Daly(2003)]{R1}
% Kopka~H and Daly~PW (2003) \textit{A Guide to \LaTeX}, 4th~edn.
% Addison-Wesley.

% \bibitem[Lamport(1994)]{R2}
% Lamport~L (1994) \textit{\LaTeX: a Document Preparation System},
% 2nd~edn. Addison-Wesley.

% \bibitem[Mittelbach and Goossens(2004)]{R3}
% Mittelbach~F and Goossens~M (2004) \textit{The \LaTeX\ Companion},
% 2nd~edn. Addison-Wesley.
% \end{thebibliography}
\bibliographystyle{sageh}   % or another style (e.g., ieeetr, unsrt, apalike)
\bibliography{References.bib}   % your .bib file name without the .bib extension
\clearpage
\onecolumn
\begin{appendices}

% ============ Appendix A ============
\section{Appendix: A}
\subsection{Derivation: Transformation of strains to force values}
\label{appendix:strain_measurement_process}

The applied force at the free end of a cantilever beam can be related to the strain experienced by the strain gauge using Euler–Bernoulli beam theory:
\[
\sigma = \frac{M c}{I}.
\]
For a rectangular section,
\[
I = \frac{b h^{3}}{12}.
\]
The reaction forces \(P_i\) are related to the stress by
\[
P_i = k\,\sigma\,\frac{b h^{2}}{6 L}.
\]
Using \(\sigma = E\,\epsilon\),
\[
P_i = G\,E\,\epsilon\,\frac{b h^{2}}{6 L}.
\]

\begin{figure}[!htbp]
    \centering
    \begin{subfigure}{0.48\linewidth}
        \centering
        \includegraphics[width=\linewidth]{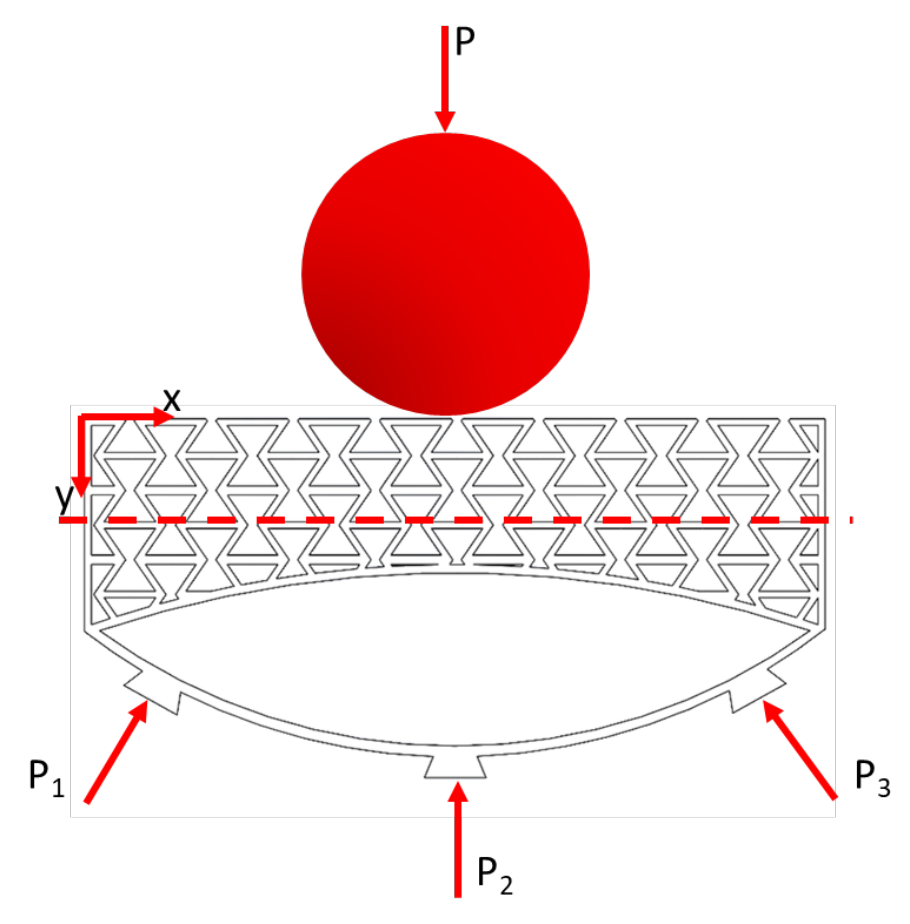}
        \caption{Figure shows the auxetic structure under the loading condition when the tomato interacts with it along with the reaction forces it encounter through the three connecting ports}
        \label{fig:Tomato_interaction_reaction forces}
    \end{subfigure}\hfill
    \begin{subfigure}{0.48\linewidth}
        \centering
        \includegraphics[width=\linewidth]{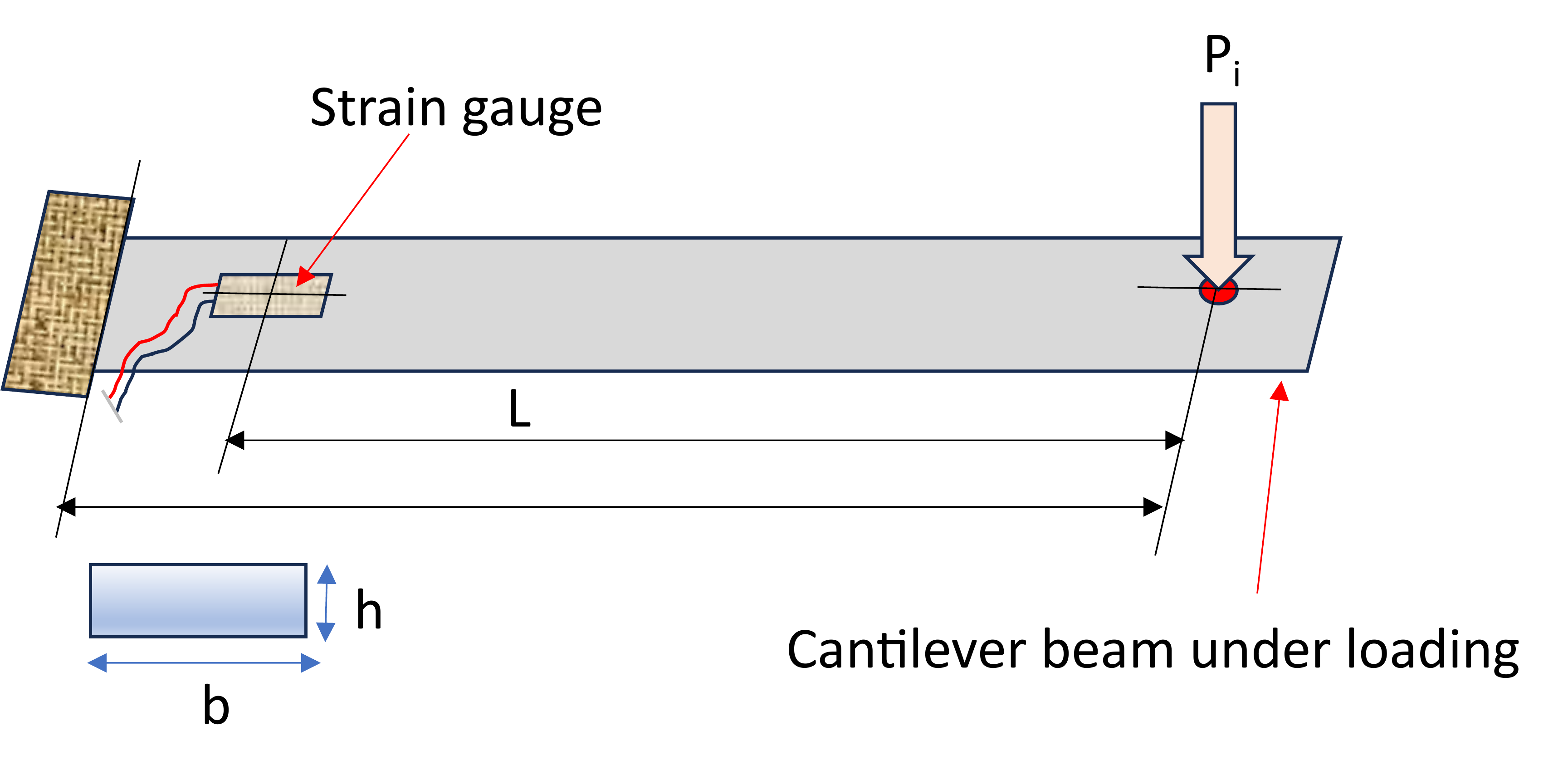}
        \caption{Figure shows the cantilever beam to measure the load at the tip through strain gauge attached at one of its section.}
        \label{fig:Cantilver_beam_strainggauge}
    \end{subfigure}
    \caption{Shows the strain measurement procedure to determine the reaction forces at the three ports of the auxetic structure experienced through rigid fingers during the interaction with the tomato on its surface}
    \label{fig:main}
\end{figure}

\noindent\textbf{Parameters used}
\begin{itemize}
    \item \(P_i\): Reaction force at port \(i\) (\(i=1,2,3\)).
    \item \(G=2.1\): Gauge factor.
    \item \(E=\qty{3.30e9}{\pascal}\): Modulus of elasticity.
    \item \(b=\qty{10.4}{\milli\metre}\), \(h=\qty{1.57}{\milli\metre}\), \(L=\qty{8}{\milli\metre}\): Beam dimensions.
    \item \(\epsilon\): Measured strain (strain gauge).
    \item Beam material: Acrylic.
\end{itemize}

% ============ Appendix B ============
\section{Appendix: B}
\subsection{Supplementary Tables}
\label{app:supp_tables}

% Single-column comparison table (no * needed in one-column mode)
\begin{table}[!htbp]
\small\sf\centering
\caption{Material properties: TPU 95A vs.\ PLA filament.}
\label{tab:material_properties}
\setlength{\tabcolsep}{6pt}
\begin{tabular}{@{}p{0.32\linewidth} p{0.31\linewidth} p{0.31\linewidth}@{}}
\toprule
\textbf{Property} & \textbf{Ultimaker TPU 95A} & \textbf{Orange PLA+ filament} \\
\midrule
Tensile strength        & 48–67     & 50–70 \\
Elongation at break     & 580\%     & 5–10\% \\
Flexural strength       & 70.5      & 80–130 \\
Flexural modulus        & 67        & 2.5–3.5 \\
Impact strength         & High      & Moderate \\
Compression strength    & 40–50     & 50–80 \\
Hardness                & 96 Shore A & 50–60 Shore D \\
Friction coefficient    & 0.3–0.6   & 0.4–0.5 \\
Fatigue strength        & Good      & Moderate \\
Print temperature       & 210–230   & 190–220 \\
Warping                 & Moderate  & Low \\
Chemical resistance     & Moderate  & Good \\
Biodegradability        & No        & Yes \\
Cost                    & Moderate  & Low \\
\bottomrule
\end{tabular}
\end{table}

\begin{table}[!htbp]
  \centering
  \caption{Specification of strain gauge GFLA-3-350-50.}
  \label{tab:strain_gauge_specification}
  \begin{tabular}{@{}p{0.42\linewidth} p{0.52\linewidth}@{}}
    \toprule
    \textbf{Specification} & \textbf{Value} \\
    \midrule
    Type & GFLA-3-350-50 \\
    Gauge length & \qty{3}{\milli\metre} \\
    Gauge resistance & \qty{350 \pm 1.0}{\ohm} \\
    Temperature compensation & for \qty{50e-6}{\per\celsius} \\
    Quantity & 10 \\
    Transverse sensitivity & \qty{1.0}{\percent} \\
    Lot No. & H50241A \\
    Batch No. & NE31K \\
    Gauge factor & \num{2.10} \(\pm\) \qty{1}{\percent} \\
    Test condition & \qty{23}{\celsius}, \qty{50}{\percent} RH \\
    \bottomrule
  \end{tabular}
\end{table}

\end{appendices}
\twocolumn

\end{document}